\documentclass{article}
\usepackage[main, final]{neurips_2026}

\usepackage[utf8]{inputenc} 
\usepackage[T1]{fontenc}    
\usepackage{hyperref}       
\usepackage{url}            
\usepackage{booktabs}       
\usepackage{amsfonts}       
\usepackage{nicefrac}       
\usepackage{microtype}      
\usepackage{comment}
\usepackage{graphicx}
\usepackage{booktabs}
\usepackage{multirow}
\usepackage{makecell}
\usepackage[table]{xcolor}
\usepackage{amsmath} 
\usepackage{wrapfig}
\usepackage{adjustbox}
\usepackage{caption}
\usepackage{float}
\usepackage{fontawesome5}

\title{CodeQuant2: Enhanced Outlier Mitigation and Adaptive Clustering for Low-Precision LRMs}

\author{%
\normalfont
\begin{tabular}{c}
\textbf{Xingyu Liu}\textsuperscript{1}\thanks{Equal contribution.}\quad
\textbf{Xiangyang Yin}\textsuperscript{2}\footnotemark[1]\quad
\textbf{Tianhua Xia}\textsuperscript{1}\quad
\textbf{Haiyu Wang}\textsuperscript{1}\quad
\textbf{Sai Qian Zhang}\textsuperscript{1,2} \\[2pt]
\textsuperscript{1}Tandon School of Engineering\\
\textsuperscript{2}Courant Institute School of Mathematics, Computing, and Data Science \\
New York University\\
\texttt{\{xl5444, shawn.yin, tx856, hw3689, sai.zhang\}@nyu.edu} \\
\href{https://github.com/SAI-Lab-NYU/CodeQuant2}{\faGithub~https://github.com/SAI-Lab-NYU/CodeQuant2}
\end{tabular}%
}

\begin{document}
\maketitle

\begin{abstract}
Large reasoning models (LRMs) improve task performance by generating longer intermediate reasoning traces, but this substantially increases inference costs and makes low-precision quantization increasingly important. However, aggressively quantized models often perform poorly on reasoning tasks. We observe larger activation fluctuations during the first decoding steps of reasoning. Calibration that omits these early states may miss their activation fluctuations. To address these reasoning-specific quantization challenges, we propose \textbf{CodeQuant2}, a clustering-based LUT quantization framework for low-precision large reasoning models. CodeQuant2 mitigates activation outliers with rotation and scaling, uses clustering to represent non-uniform weight distributions, and adapts codebook sizes to the quantization difficulty of each layer. Guided by our observation of reasoning-time activation behavior, CodeQuant2 further improves calibration sample selection to better match actual decoding statistics. Combined with our customized LUT CUDA kernel, CodeQuant2 delivers substantial accuracy improvements on reasoning benchmarks while achieving up to $2.85\times$ speedup over BF16.
\end{abstract}

\section{Introduction}
Large reasoning models (LRMs)~\cite{guo2025deepseek,yang2025qwen3,agarwal2025gpt} have become important for structured, multi-step tasks such as mathematical reasoning, code generation, and tool-augmented decision making. By spending more computation on reasoning, these models achieve stronger performance but incur higher inference costs from longer generation traces, larger memory footprints, and increased compute demand. Low-precision quantization is therefore a key strategy for efficient deployment. Recent hardware support further strengthens this trend: NVIDIA Hopper~\citep{nvidia_h100_whitepaper} and Ada GPUs~\citep{nvidia_ada_gpu_arch} support FP8 arithmetic, while Blackwell extends it to FP4~\cite{nvfp4}, making low-precision LRM inference increasingly practical.

However, activation outliers remain a major obstacle to low-bit quantization in LRMs, as in conventional foundation models~\cite{dettmers2022llmint88bitmatrixmultiplication,sun2024massiveactivationslargelanguage}. These extreme values enlarge the quantization range, leaving limited effective precision for most activations and causing large approximation errors. In reasoning generation, such errors are even more harmful because they can accumulate across autoregressive decoding steps~\cite{liu2025quantization,li2025quantizationmeetsreasoningexploring,liang2025paroquant}. Since reasoning tasks often involve long multi-step traces, errors introduced early can be amplified later and degrade final reasoning quality~\cite{zhao2025qspec,liu2025quantization}. Therefore, despite progress in outlier-smoothing post-training quantization methods~\cite{tseng2024quipbetterllmquantization,chee2024quip2bitquantizationlarge,ashkboos2024quarot,xiao2023smoothquant,xiang2024dfrot,liu2024spinquant,liang2025paroquant,yin2026codequant}, aggressive low-bit quantization still struggles to maintain stable performance on complex reasoning tasks.

Existing methods address these challenges from complementary directions. Outlier-smoothing techniques, including scaling and rotation, mitigate activation quantization errors~\cite{xiao2023smoothquant,ashkboos2024quarot,liu2024spinquant,liang2025paroquant}. Meanwhile, clustering-based quantization represents weights using learned centroids, better capturing non-uniform weight distributions~\cite{kim2023squeezellm,yin2026codequant}. Moreover, clustering-based representations naturally support lookup-table (LUT) inference, offering opportunities for efficient low-bit deployment. Building on these complementary approaches, we adopt clustering-based LUT quantization as the foundation of our framework and identify two key considerations for accurate and efficient inference.

Although existing outlier-mitigation techniques improve activation quantization, these gains do not always translate into comparable improvements on reasoning tasks. To understand this gap, we examine activation behavior during reasoning and observe higher activation fluctuations and channel-wise outliers in early decoding steps, which gradually weaken as generation progresses. Existing outlier-mitigation techniques do not naturally account for these reasoning-specific decoding-stage-dependent activation patterns. \textbf{Thus, we argue that outlier mitigation should explicitly emphasize high-variance early-decoding activations to better preserve reasoning quality.} This motivates us to design a calibration strategy that better captures these early-stage activation patterns.

Beyond activation calibration, clustering-based quantization offers another degree of freedom: the number of clusters assigned to each weight matrix. Allocating more clusters to sensitive matrices can improve quantization quality, but also increases LUT inference latency. \textbf{Thus, effective cluster allocation should balance quantization performance and hardware-aware inference latency, rather than optimizing accuracy alone.} We therefore design a performance--latency-balanced cluster allocation strategy based on matrix sensitivity and profiled execution costs.

Building on these two design principles, we present \textit{CodeQuant2}, a clustering-based low-bit quantization framework for LRMs. Our contributions can be summarized as follows:
\begin{itemize}
    \item We propose \textit{CodeQuant2}, a clustering-based quantization framework that combines reasoning-aware calibration, fine-tuned rotation and scaling transformations, and differentiable assignment tuning, enabling learned cluster assignments to handle outlier-heavy distributions.
    \item We further introduce \textit{Adaptive Cluster Allocation} (ACA), which adaptively assigns cluster numbers by balancing accuracy and hardware efficiency for each target matrix.
    \item We design a GPU kernel that directly executes on the clustered representation by recasting decode-time matrix multiplication as table-lookup computation. In the evaluated GPU settings, CodeQuant2 achieves up to \(2.85\times\) speedup while improving reasoning accuracy.
\end{itemize}

\section{Related Work}
\subsection{Large Reasoning Model}
Chain-of-thought prompting and related methods show that explicitly unfolding intermediate reasoning steps can improve performance on complex tasks~\cite{wei2022chain,wang2022self,zelikman2024star}. Recent large reasoning models internalize this capability through longer reasoning traces or dedicated thinking modes, enabled by additional test-time compute~\cite{shao2024deepseekmath,guo2025deepseek,yang2025qwen3,agarwal2025gpt}. Their training increasingly relies on reinforcement-learning-based post-training, from earlier PPO-style methods~\cite{luong2024reft,xie2025interleaved} to recent GRPO-based approaches and variants~\cite{shao2024deepseekmath,guo2025deepseek,yu2025dapo,gao2025soft}. Thus, explicit reasoning traces, extra test-time computation, and RL-driven post-training have become defining features of current LRMs. Despite their strong reasoning ability, LRMs are more fragile under aggressive low-bit quantization than standard language models. Prior work shows that quantization can cause much larger degradation on reasoning benchmarks, even when general-task performance remains relatively stable~\cite{liu2025quantization,li2025quantizationmeetsreasoningexploring,zhao2025qspec}. This is largely because quantization errors during autoregressive generation can accumulate over long reasoning traces, making low-bit LRM quantization more challenging than quantizing conventional foundation models~\cite{liu2025quantization,liang2025paroquant,zhao2025qspec}.

\subsection{Outlier Aware Quantization}
Activation outliers are a major challenge for low-bit LLM quantization. Prior work has shown that the input activations to self-attention weight matrices often exhibit pronounced~\textit{channel-wise outliers}~\cite{dettmers2022gpt3,sun2024massive,xiang2024dfrot}, where a small number of channels have much larger magnitudes than the rest. These outliers dominate the quantization range and introduce large quantization errors when low-bit quantization is applied.
One solution is to isolate or retain outliers in higher precision, as in mixed-precision methods such as LLM.int8()~\cite{dettmers2022gpt3} and SPQR~\cite{dettmers2023spqr}. While effective, these methods often introduce irregular computation and memory access, reducing hardware efficiency.

To preserve uniform low-precision execution, many methods instead apply computational invariant transformations before quantization. Scaling-based methods, such as SmoothQuant~\cite{xiao2023smoothquant}, use channel-wise scaling to redistribute quantization difficulty between activations and weights, while rotation-based methods, such as QuaRot~\cite{ashkboos2024quarot}, SpinQuant~\cite{liu2024spinquant}, and ParoQuant~\cite{liang2025paroquant}, spread outliers through orthogonal transformations. Let $\mathbf{X}\in\mathbb{R}^{N\times d_{\mathrm{in}}}$ be the hidden states and $\mathbf{T}\in\mathbb{R}^{d_{\mathrm{in}}\times d_{\mathrm{in}}}$ an invertible transformation along the hidden dimension. For a linear projection $\mathbf{W}\in\mathbb{R}^{d_{\mathrm{out}}\times d_{\mathrm{in}}}$, inserting $\mathbf{T}\mathbf{T}^{-1}$ preserves the output given that $\mathbf{X}\mathbf{W}^{\top}=\mathbf{X}\mathbf{T}\mathbf{T}^{-1}\mathbf{W}^{\top}=(\mathbf{X}\mathbf{T})(\mathbf{W}\mathbf{T}^{-\top})^{\top}$.

The transformation $\mathbf{T}$ can be instantiated as an orthogonal rotation or a diagonal scaling matrix. These transformations reduce outlier concentration and improve robustness under low-precision quantization. More recently, CodeQuant~\cite{yin2026codequant} combines rotation with clustering-based non-uniform quantization. However, aggressive low-bit quantization remains challenging, particularly for reasoning tasks.

\subsection{Non-uniform Quantization}
Non-uniform quantization provides a more flexible alternative to fixed uniform bins for the highly irregular distributions encountered in aggressive low-bit LLM compression. Existing work in this direction can be roughly divided into two categories. The first adjusts scalar quantization levels to better match the underlying distribution, as exemplified by NF4 in QLoRA~\cite{dettmers2023qlora}. The second adopts stronger clustering-based representations to improve expressiveness in the low-bit regime, as explored in SqueezeLLM and CodeQuant~\cite{kim2023squeezellm,yin2026codequant}. Related work has also investigated vector quantization as an even stronger low-bit representation~\cite{liu2024vptq}. Overall, non-uniform quantization has shown clear advantages over uniform quantization, but existing methods still focus on static reconstruction quality and general compression settings, leaving low-bit quantization for LRMs underexplored.

\subsection{GEMM with Lookup Table-based Implementation}
\label{sec:bg:LUT}
GEMM with clustered multiplicands requires efficient Lookup Table (LUT) support for practical deployment. Without hardware-friendly LUTs, centroids must be stored and loaded as floating-point values during computation, introducing substantial memory and execution overhead. Existing work has approached this problem from both CPU and GPU settings. On CPUs, DeepGEMM~\citep{ganji2023deepgemm}, LUTIN~\citep{lin2024lutin}, and T-MAC~\citep{wei2025t} explore LUT-oriented execution paths for low-precision inference, with an emphasis on kernel design, memory efficiency, and reformulating mixed-precision GEMM as table lookup. On GPUs, LUT-GEMM~\citep{park2022lut}, FLUTE~\citep{guo2024fast}, and LUT Tensor Core~\citep{mo2025lut} develop increasingly specialized kernels and software-hardware co-design techniques to reduce lookup overhead and better integrate LUT computation into high-throughput matrix multiplication pipelines.

\section{Methodology}
\label{sec:methdology}
\begin{figure}[t]
    \centering
    \includegraphics[width=1.0\textwidth]{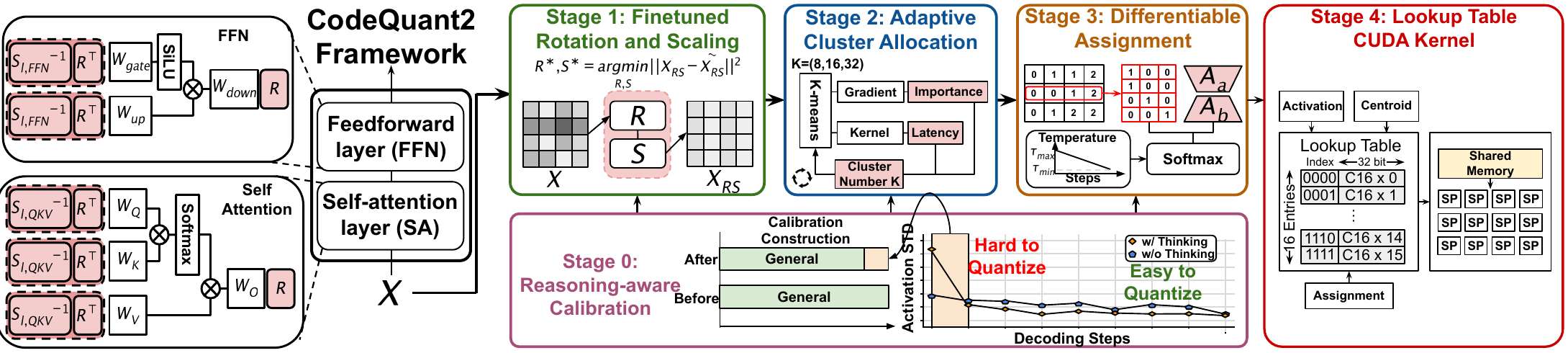}
    \caption{
    Overview of the CodeQuant2 framework. \textit{Left:} Rotation and scaling are fused into the model for outlier smoothing. \textit{Right:} CodeQuant2 applies fine-tuned rotation and scaling, allocates clusters adaptively, tunes centroids and assignments, and uses a LUT-based kernel for inference.}
    \label{fig:framework}
    \par\vspace{-10pt}
\end{figure}
In this section, we present the methodology of \textbf{CodeQuant2}, as shown in Figure~\ref{fig:framework}. We first motivate the design by showing that  decoding on reasoning tasks induces activation behaviors not well captured by conventional calibration data. We then introduce CodeQuant2’s main components, including fine-tuned rotation and scaling for outlier mitigation, adaptive cluster allocation, and differentiable clustering quantization. Finally, we describe the LUT-based GPU kernel for efficient inference.

\subsection{Outlier Behavior in Large Reasoning Models}
\label{sec:lrm-data-dist}
\begin{wrapfigure}{r}{0.60\textwidth}
    \centering
    \vspace{-1.0em}
\includegraphics[width=0.60\textwidth]{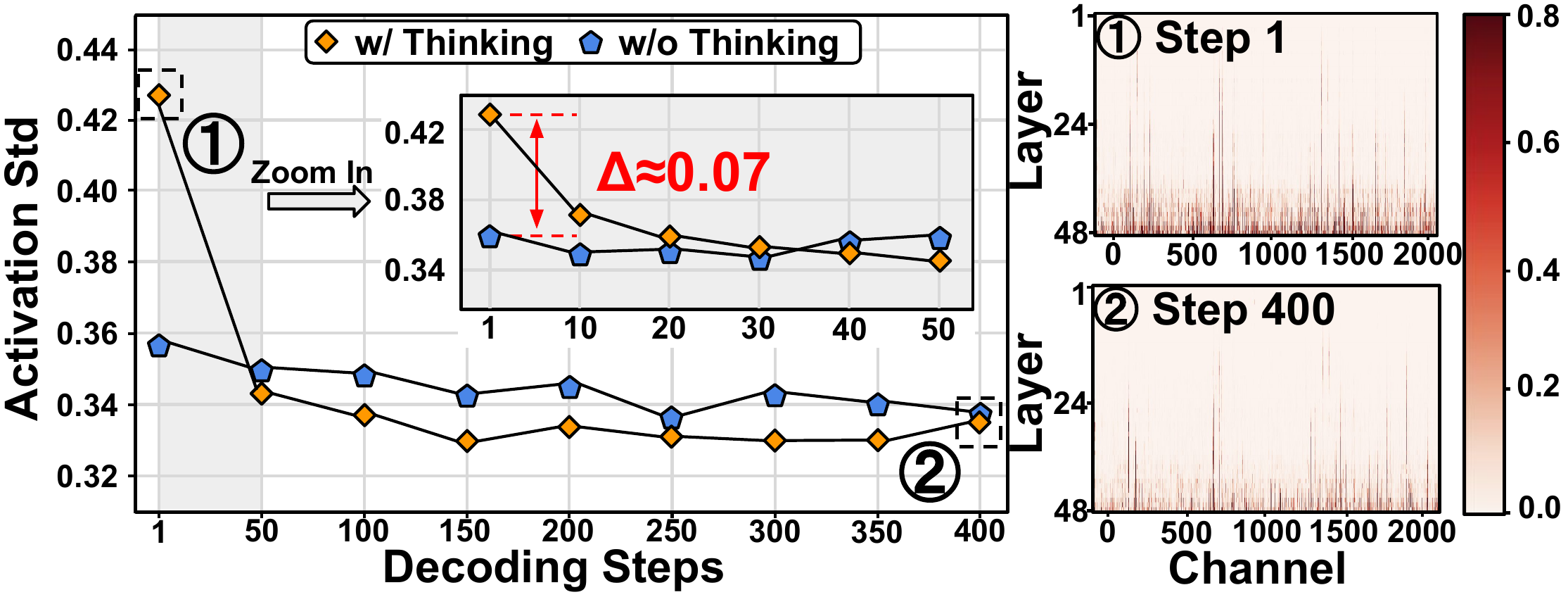}
    \vspace{-1.0em}
    \caption{
    Activation behavior during reasoning decoding on GSM8K~\cite{cobbe2021training}, measured on Qwen3-4B-Thinking~\cite{qwen3technicalreport} and averaged over 128 samples.
 \textit{Left:} Averaged activation standard deviation across decoding steps with and without thinking. \textit{Right:} Activation heatmaps at representative decoding steps.
    }
    \label{fig:calibration_motivation}
    \par\vspace{-10pt}
\end{wrapfigure}

To better characterize outlier behavior in LRMs, we analyze the distributions of self-attention activations on reasoning benchmarks, including GSM8K~\cite{cobbe2021training} and AIME25~\cite{aime25}. As shown in Figure~\ref{fig:calibration_motivation}, models generating reasoning traces exhibit substantially larger activation fluctuations during the early decoding stage. In particular, the average activation standard deviation over the first 30 decoding steps is about $20\%$ higher than that of non-thinking decoding, while non-thinking statistics remain relatively stable across steps. The heatmaps on the right of Figure~\ref{fig:calibration_motivation} show the channel-wise activation standard deviation across layers at decoding steps 1 and 400 for LRMs. At step 1, strong channel-wise outliers appear from shallow layers and persist toward deeper layers. By contrast, at later decoding steps, the outlier pattern becomes noticeably weaker.

Because quantization errors introduced early in reasoning can propagate through later decoding steps and affect subsequent token generation, reducing error at the beginning of the reasoning process is especially important. This observation motivates our \textit{reasoning-aware calibration} strategy. During local fine-tuning, we combine activations from early tokens decoded by the full-precision model with general-text samples. Section~\ref{sec:experiment} gives the sampling details. Additional results across reasoning models and datasets appear in Appendix~\ref{sec:app:activation_fluctuation}.

\subsection{Fine-tuned Rotation and Scaling for Outlier Smoothing}
\label{sec:rotation-scaling}
Motivated by the outlier patterns in Figure~\ref{fig:calibration_motivation}, we aim to smooth input activations further before low-precision quantization. To this end, we jointly apply rotation and scaling for outlier mitigation. Specifically, we introduce a shared orthogonal rotation matrix $\mathbf{R}$ and layer-wise diagonal scaling matrices $\mathbf{S}_{\ell,t}$, where $\ell$ denotes the layer index and $t\in \{{\mathrm{QKV},\mathrm{FFN}}\}$ denotes the projection group. We apply rotation, scaling, and quantization only to the query, key, value, gate, and up projections. Given an input activation $\mathbf{X}$ and its corresponding weight matrix $\mathbf{W}\in\mathbb{R}^{d_{\mathrm{out}}\times d_{\mathrm{in}}}$, the transformed activation and weight are defined as $\mathbf{X}_{RS}=\mathbf{X}\mathbf{R}\mathbf{S}_{\ell,t}$ and $\mathbf{W}_{RS}=\mathbf{W}\mathbf{R}\mathbf{S}_{\ell,t}^{-1}$, respectively. This transformation preserves the original linear operation since $\mathbf{X}_{RS}\mathbf{W}_{RS}^{\top}=\mathbf{X}\mathbf{W}^{\top}$. The transformed weights can be computed offline. We optimize $\mathbf{R}$ and $\mathbf{S}_{\ell,t}$ through local fine-tuning by minimizing the activation quantization error over the calibration set:
\begin{equation}
    \label{eqn:activation-optimization}
    \mathbf{R}^{\star}, \{\mathbf{S}_{\ell,t}^{\star}\}
    =
    \operatorname*{arg\,min}_{\mathbf{R}, \{\mathbf{S}_{\ell,t}\}}
    \left\|
    \mathbf{X}_{RS}
    -
    \widetilde{\mathbf{X}}_{RS}
    \right\|^{2}
\end{equation}
where $\widetilde{\mathbf{X}}_{RS}$ denotes the quantized activation. We parameterize $\mathbf{R}$ with the Cayley transform~\citep{nishimori2005learning,li2020efficient,liu2024spinquant} to preserve orthogonality during optimization.

\subsection{Adaptive Cluster Allocation}
\label{sec:aca}
After the input activations are transformed as described in Section~\ref{sec:rotation-scaling}, activation quantization error is reduced, but preserving the full-precision layer output still requires careful weight clustering. In CodeQuant2, each weight value is mapped to one of the $K$ centroids in a learned codebook, with details described in Section~\ref{sec:clustering_fine-tuning}. Therefore, the cluster number $K$ directly determines the codebook size and directly affects both quantization accuracy and system latency. Since $K=2^b$ corresponds naturally to a $b$-bit index, existing clustering-based methods typically choose $K$ based on the target bitwidth and use the same cluster number for all target linear layers~\cite{kim2023squeezellm,yin2026codequant}. However, such uniform assignment is often suboptimal. Empirically, matrices exhibit different sensitivities to clustering. Some can be represented with fewer clusters while introducing limited output perturbation, whereas others require more clusters to preserve accuracy.

Motivated by the effectiveness of mixed precision, we propose to adaptively assign the cluster number $K$ to balance matrix importance and hardware performance. For simplicity, we adjust $K$ at the matrix level, treating it as a capacity parameter. We restrict $K$ to a predefined candidate set $\mathcal{K}$.
In our experiments, $\mathcal{K}=\{8,16,32\}$. We initialize all target matrices with the original cluster number $K_{\mathrm{ori}}$, and then estimate the effect of changing $K$ for each matrix before selecting which matrices to update. 

The impact of cluster-number adjustment on accuracy is estimated through the local reconstruction error. Given the original cluster number $K_{\mathrm{ori}}$ and the corresponding clustered weight $\widetilde{\mathbf{W}}_{RS}^{(K_{\mathrm{ori}})}$, the clustered output $\widetilde{\mathbf{Y}}^{(K_{\mathrm{ori}})}$ under this cluster assignment is written as
$\widetilde{\mathbf{Y}}^{(K_{\mathrm{ori}})}
=
\widetilde{\mathbf{X}}_{RS}(\widetilde{\mathbf{W}}_{RS}^{(K_{\mathrm{ori}})})^{\top}$.
When changing the cluster number from $K_{\mathrm{ori}}$ to $K_{\mathrm{tar}}\in\mathcal{K}$, the corresponding local output perturbation is
$\Delta\mathbf{Y}_{K_{\mathrm{ori}}\rightarrow K_{\mathrm{tar}}}
=
\widetilde{\mathbf{X}}_{RS}
(
(\widetilde{\mathbf{W}}_{RS}^{(K_{\mathrm{ori}})})^{\top}
-
(\widetilde{\mathbf{W}}_{RS}^{(K_{\mathrm{tar}})})^{\top}
)$.
The full-precision output of this layer on the calibration inputs is $\mathbf{Y}_{\mathrm{fp}}=\mathbf{X}\mathbf{W}^{\top}=\mathbf{X}_{RS}\mathbf{W}_{RS}^{\top}$. To estimate the accuracy impact of this change, we use a first-order approximation of the local reconstruction loss
$\mathcal{L}_{\mathrm{recon}}=||\mathbf{Y}_{\mathrm{fp}}-\widetilde{\mathbf{Y}}^{(K_{\mathrm{ori}})}||^{2}$.
This gives the following importance score:
\begin{equation}
\label{eqn:cluster_assignment}
I_{K_{\mathrm{ori}}\rightarrow K_{\mathrm{tar}}}
=
\frac{1}{N}
\left\|
\Delta\mathbf{Y}_{K_{\mathrm{ori}}\rightarrow K_{\mathrm{tar}}}
\odot
\nabla_{\widetilde{\mathbf{Y}}}
\mathcal{L}_{\mathrm{recon}}
\right\|_F^2.
\end{equation}
where $\nabla_{\widetilde{\mathbf{Y}}}\mathcal{L}_{\mathrm{recon}}$ denotes the gradient of the local reconstruction loss with respect to the clustered output. The derivation is provided in Appendix~\ref{sec:app:importance_score}.
For efficiency, CodeQuant2 computes this score using the local reconstruction loss, although the same formulation can also be extended to the final training loss of LRMs.
This choice avoids full model-level backpropagation and only requires gradient computation for the local GEMM operation, substantially reducing the computational overhead.

Considering only the importance score leads to a trivial solution, since it would assign the maximum number of clusters to every group. On the other hand, increasing $K$ also introduces higher LUT-based processing latency. To balance quantization quality and system performance, we profile the LUT execution latency of each target matrix under different cluster numbers on the target GPU platform. Let $T_m(K)$ denote the profiled lookup latency of weight matrix $m$ when using $K$ clusters. All matrices are initialized with $K_{\mathrm{ori}}$, with latency $T_m(K_{\mathrm{ori}})$. Starting from 16 clusters (4-bit indices), we pair a change to 32 clusters (5-bit indices) for one matrix with a change to 8 clusters (3-bit indices) for another. Each pair has an average index width of 4 bits across the two matrices. We also account for their profiled latency. We increase $K$ for the matrix with the largest estimated benefit and decrease $K$ for the matrix with the smallest estimated loss, subject to the latency budget. Figure~\ref{fig:aca_and_clustering}(a) summarizes this allocation process.

\subsection{Learnable Clustering for Enhanced Performance}
\label{sec:clustering_fine-tuning}

\begin{figure}[t]
    \centering
    \includegraphics[width=\linewidth]{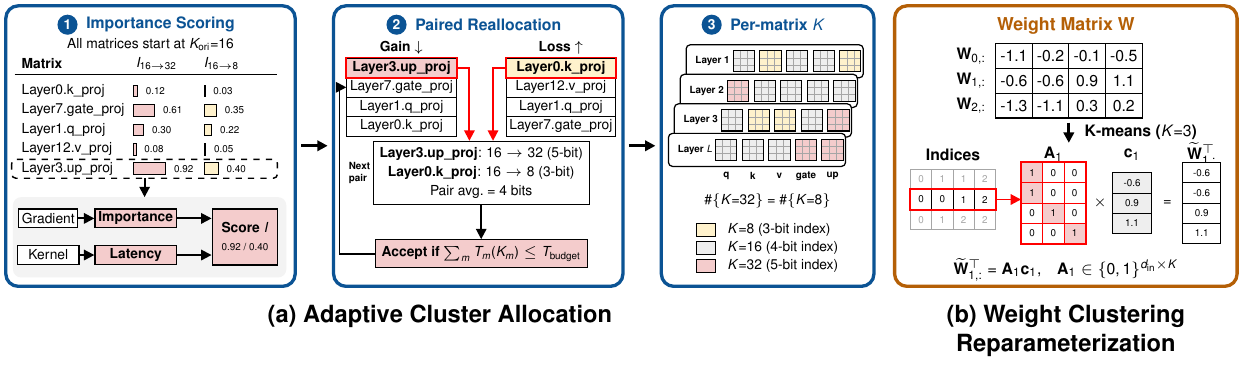}
    \caption{Adaptive cluster allocation and clustering. \textit{Left:} Importance scores and latency guide cluster allocation. \textit{Right:} K-means initializes the centroid matrix and per-row binary assignment matrices.}\label{fig:aca_and_clustering}
    \par\vspace{-10pt}
\end{figure}

To better align the quantized model with the full-precision model, CodeQuant2 fine-tunes the clustered weights by optimizing a centroid matrix and learning soft assignment matrices while keeping the initial K-means assignments fixed. Let $\mathbf{W}\in\mathbb{R}^{d_{\mathrm{out}}\times d_{\mathrm{in}}}$ be a weight matrix. Its $i$-th row is $\mathbf{W}_{i,:}$, where $0\leq i<d_{\mathrm{out}}$. Figure~\ref{fig:aca_and_clustering}(b) shows an example weight matrix $\mathbf{W}\in\mathbb{R}^{3\times 4}$ clustered with $K=3$. We initialize the clustering by applying K-means to each row. For row $i$, the clustered weight is
\begin{equation}                    
    \widetilde{\mathbf{W}}_{i,:}^{\top}=\mathbf{A}_i\mathbf{c}_i.
\end{equation}
The centroid matrix $\mathbf{C}\in\mathbb{R}^{d_{\mathrm{out}}\times K}$ stores one codebook per output row. We write its $i$-th row as $\mathbf{c}_i^{\top}$, where $\mathbf{c}_i\in\mathbb{R}^{K}$. The binary matrix $\mathbf{A}_i\in\{0,1\}^{d_{\mathrm{in}}\times K}$ records the initial K-means assignments for row $i$ of $\mathbf{W}$. Its $j$-th row selects the centroid for $W_{ij}$.
We use $\mathbf{Y}_{\mathrm{fp}}$, defined before Eq.~\ref{eqn:cluster_assignment}, as the reference output for the local reconstruction loss:
\begin{equation}
\mathcal{L}_{\mathrm{recon}}=\left\|\mathbf{Y}_{\mathrm{fp}}-\widetilde{\mathbf{X}}_{RS}\widetilde{\mathbf{W}}_{RS}^{\top}\right\|_F^2.
\end{equation}

During fine-tuning, we keep the initial K-means assignments fixed and learn soft assignments through low-rank updates. We stack the one-hot assignment matrices $\mathbf{A}_i$ into $\mathbf{A}\in\{0,1\}^{d_{\mathrm{out}}\times d_{\mathrm{in}}\times K}$. To reduce the number of trainable parameters, we apply LoRA to the first two dimensions:
\begin{equation}
    \mathbf{A}' = \operatorname{softmax}\left(
        \frac{\mathbf{A}+\mathbf{B}^d\mathbf{B}^u}{\tau}
    \right),
\end{equation}
where $\mathbf{B}^d\in\mathbb{R}^{d_{\mathrm{out}}\times r\times K}$ and $\mathbf{B}^u\in\mathbb{R}^{r\times d_{\mathrm{in}}\times K}$ are the low-rank factors, with $r\ll\min(d_{\mathrm{out}},d_{\mathrm{in}})$. Their multiplication is performed independently for each centroid, while the softmax operates along the centroid dimension. The resulting assignments reconstruct each weight row as $\widetilde{\mathbf{W}}_{i,:}^{\top}=\mathbf{A}'_i\mathbf{c}_i$.

The temperature $\tau$ controls the sharpness of the soft assignments and is annealed during fine-tuning:
\begin{equation}
    \tau(t)=\tau_{\max}-\frac{t}{T}(\tau_{\max}-\tau_{\min}),
    \quad t=0,\ldots,T,
\end{equation}
where $t$ denotes the current fine-tuning step, $T$ is the total number of steps, and $\tau_{\max}$ and $\tau_{\min}$ are the initial and final temperatures, respectively. During the final stage of fine-tuning, we convert soft assignments to hard assignments, allowing the centroids to adapt to the discrete assignments.

\subsection{Lookup Table CUDA Kernel Design}
\label{sec:hardware_implementation}
\begin{wrapfigure}{r}{0.65\textwidth}
    \centering
    \vspace{-1.0em}
\includegraphics[width=0.65\textwidth]{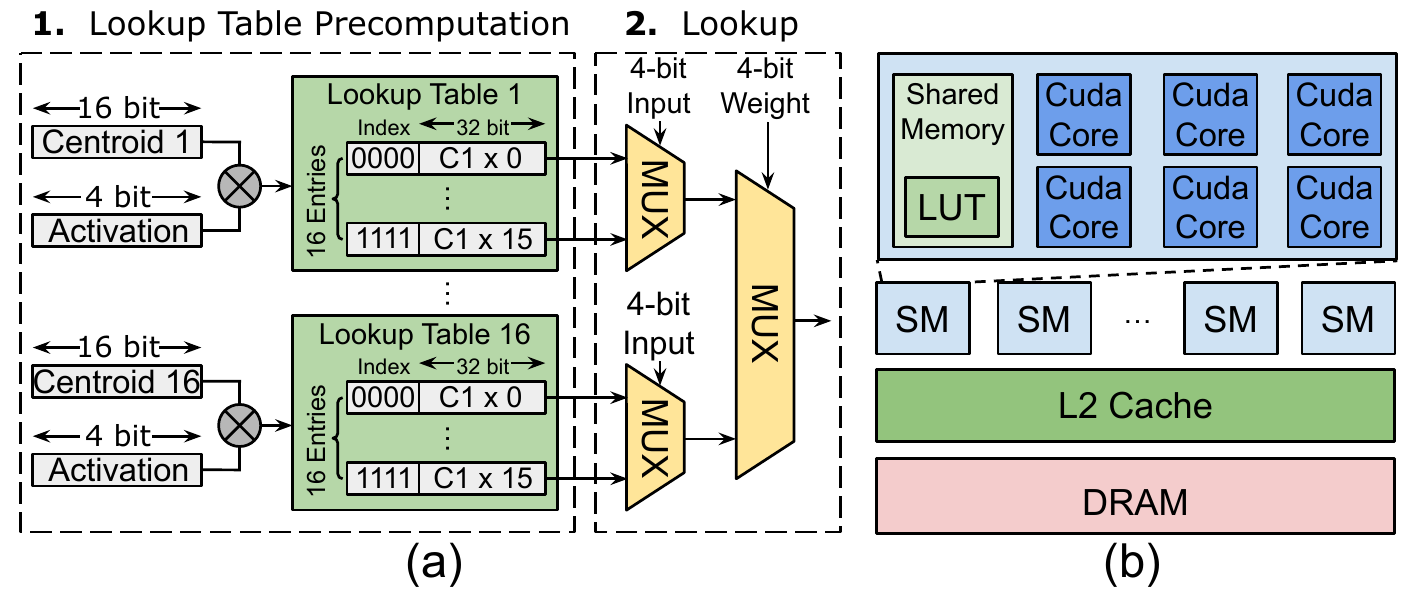}
    \vspace{-1.0em}
    \caption{(a) The steps of the LUT kernel, including a one-time LUT precomputation and table lookup. (b) The precomputed LUT is stored in the shared memory in the Streaming Multiprocessors.
    }
    \label{fig:lut_hw}
    \vspace{-1.0em}
\end{wrapfigure}
CodeQuant2 applies linear quantization to input activations and clustering-based quantization to weights. A simple deployment strategy is to reconstruct the clustered weight matrix from centroid indices and then invoke a standard CUDA GEMM kernel~\cite{kim2023squeezellm}. However, this reconstruction-based approach introduces additional dequantization and memory traffic before matrix multiplication, reducing the practical benefit of the low-bit weight format. To avoid this overhead, we design a GPU kernel that directly operates on the clustered representation by reformulating decode-time matrix multiplication as a table-lookup computation.

An example kernel design is illustrated in Figure~\ref{fig:lut_hw}(a), where we consider 4-bit activation quantization and weight clustering with 16 centroids. The same principle can be extended to other bit-widths for activation and codebook sizes for weight clustering. In step 1 in Figure~\ref{fig:lut_hw}(a), with 4-bit activations, each activation element is mapped to one of 16 quantized levels, while each weight element is stored as a 4-bit assignment index selecting one centroid from the row-wise codebook. The key idea is to precompute the products between quantized activation levels and weight centroids once, so that each multiplication between a quantized activation and a clustered weight can be replaced by a table lookup indexed by the activation level and the weight assignment, as shown in Step 2 of Figure~\ref{fig:lut_hw}(a).

We further improve the GPU kernel for decode-time execution with three optimizations. First, activation min/max computation, 4-bit quantization, and bit packing are fused, so the activation vector is read only once and directly converted into packed indices consumed by the kernel. This avoids writing intermediate quantized activations to global memory. Second, the LUT is stored in GPU shared memory within each streaming multiprocessor (SM) and reused during decoding, as illustrated in Figure~\ref{fig:lut_hw}(b), reducing dequantization overhead and improving execution efficiency. Third, because decode-time GEMM exposes limited batch-level parallelism, we tune the matrix tile sizes to better match CUDA core resources and improve hardware utilization. Together, these optimizations enable clustered low-bit models to execute efficiently through table lookup, avoiding the higher latency of reconstruction-based matrix multiplication.

\begin{table}[!t]
\caption{Performance under A4W4 setting in terms of perplexity on WikiText2 and C4, and accuracy on Math500 (MA500), GSM8K, AIME25, AIME26, HumanEval (HE), and LiveCodeBench (LCB). }
\centering
\resizebox{1\columnwidth}{!}{
\begin{tabular}{c|c|cc|cccccc|c}
    \toprule
    
    \textbf{Models} & \textbf{Methods} 
    & \textbf{Wiki2} ($\downarrow$) 
    & \textbf{C4} ($\downarrow$) 
    & \textbf{MA500} ($\uparrow$) 
    & \textbf{GSM8K} ($\uparrow$) 
    & \textbf{AIME25} ($\uparrow$) 
    & \textbf{AIME26} ($\uparrow$) 
    & \textbf{HE} ($\uparrow$) 
    & \textbf{LCB} ($\uparrow$) 
    & \textbf{Avg} ($\uparrow$) \\
    
    \midrule

    \multirow{6}{*}{\makecell{Qwen3-\\4B-Thinking}} 
    & BF16       & 11.478 & 18.391 & 0.958 & 0.952 & 0.688 & 0.708 & 0.963 & 0.758 & 0.838 \\
    & QuaRot     & 17.366 & 26.879 & 0.394 & 0.506 & 0.004 & 0.000 & 0.000 & 0.002 & 0.151 \\
    & ParoQuant  & 14.771 & 23.070 & 0.720 & 0.847 & 0.183 & 0.142 & 0.549 & 0.180 & 0.437 \\
    & SqueezeLLM & 14.627 & 22.642 & 0.852 & 0.908 & 0.192 & 0.158 & 0.695 & 0.158 & 0.494 \\
    & CodeQuant  & 13.623 & 21.410 & 0.930 & 0.929 & 0.392 & 0.367 & 0.902 & 0.464 & 0.664 \\
    & \cellcolor{blue!10}\textbf{Ours}
                 & \cellcolor{blue!10}\textbf{12.890}
                 & \cellcolor{blue!10}\textbf{20.538}
                 & \cellcolor{blue!10}\textbf{0.950}
                 & \cellcolor{blue!10}\textbf{0.938}
                 & \cellcolor{blue!10}\textbf{0.563}
                 & \cellcolor{blue!10}\textbf{0.613}
                 & \cellcolor{blue!10}\textbf{0.945}
                 & \cellcolor{blue!10}\textbf{0.614}
                 & \cellcolor{blue!10}\textbf{0.770} \\
    
    \cmidrule(lr){1-11}
    
    \multirow{6}{*}{\makecell{Qwen3-\\8B}} 
    & BF16       & 10.047 & 15.439 & 0.948 & 0.958 & 0.625 & 0.646 & 0.921 & 0.732 & 0.805 \\
    & QuaRot     & 14.475 & 21.059 & 0.382 & 0.259 & 0.033 & 0.004 & 0.037 & 0.003 & 0.120 \\
    & ParoQuant  & 12.830 & 19.260 & 0.762 & 0.895 & 0.196 & 0.167 & 0.450 & 0.182 & 0.442 \\
    & SqueezeLLM & 12.215 & 18.515 & 0.850 & 0.900 & 0.242 & 0.179 & 0.805 & 0.294 & 0.545 \\
    & CodeQuant  & \textbf{10.509} & \textbf{16.349} & 0.934 & \textbf{0.946} & 0.446 & 0.483 & 0.895 & \textbf{0.589} & 0.716 \\
    & \cellcolor{blue!10}\textbf{Ours}
                 & \cellcolor{blue!10}11.362
                 & \cellcolor{blue!10}17.177
                 & \cellcolor{blue!10}\textbf{0.936}
                 & \cellcolor{blue!10}0.945
                 & \cellcolor{blue!10}\textbf{0.479}
                 & \cellcolor{blue!10}\textbf{0.504}
                 & \cellcolor{blue!10}\textbf{0.915}
                 & \cellcolor{blue!10}0.566
                 & \cellcolor{blue!10}\textbf{0.724} \\

    \cmidrule(lr){1-11}
    
    \multirow{6}{*}{\makecell{Phi-4-\\Reasoning}}
& BF16 & 7.393 & 13.187 & 0.858 & 0.943 & 0.283 & 0.250 & 0.872 & 0.410 & 0.603 \\
& QuaRot & 9.511 & 16.564 & 0.766 & 0.908 & 0.192 & 0.129 & 0.640 & 0.238 & 0.479 \\
& ParoQuant & 9.489 & 16.584 & 0.770 & 0.908 & 0.183 & 0.150 & 0.683 & 0.235 & 0.488 \\
& SqueezeLLM & 8.375 & 14.871 & 0.806 & 0.921 & \textbf{0.258} & 0.188 & 0.750 & 0.321 & 0.541 \\
& CodeQuant & 8.087 & 14.358 & \textbf{0.840} & \textbf{0.940} & 0.246 & 0.242 & 0.732 & 0.353 & 0.559 \\
& \cellcolor{blue!10}\textbf{Ours}
& \cellcolor{blue!10}\textbf{7.739}
& \cellcolor{blue!10}\textbf{13.660}
& \cellcolor{blue!10}\textbf{0.840}
& \cellcolor{blue!10}0.932
& \cellcolor{blue!10}0.246
& \cellcolor{blue!10}\textbf{0.250}
& \cellcolor{blue!10}\textbf{0.823}
& \cellcolor{blue!10}\textbf{0.362}
& \cellcolor{blue!10}\textbf{0.575} \\
    
    \bottomrule
            
\end{tabular}
}
\label{tab:eval:main}
\par\vspace{-10pt}
\end{table}

\section{Experiment}
\label{sec:experiment}
We evaluate our method on LRMs of different scales, including Qwen3-4B-Thinking~\citep{yang2025qwen3}, Qwen3-8B with thinking mode enabled~\citep{yang2025qwen3}, and Phi-4-Reasoning~\citep{abdin2025phi4reasoningtechnicalreport}. Our evaluation covers language modeling, general question answering, mathematical reasoning, and code generation. For language modeling, we report perplexity on WikiText2~\citep{merity2016pointersentinelmixturemodels} and C4~\citep{raffel2023exploringlimitstransferlearning}. For general question answering, we report ARC~\citep{clark2018thinksolvedquestionanswering}, PIQA~\citep{bisk2020piqa}, and WinoGrande~\citep{sakaguchi2021winogrande} results in the appendix. For mathematical reasoning, we evaluate on MATH500~\citep{hendrycks2021measuringmassivemultitasklanguage}, GSM8K~\citep{cobbe2021trainingverifierssolvemath}, and AIME~\citep{aime25,aime26} under both pass@1 and pass@8. To assess code generation after quantization, we use HumanEval~\citep{chen2021evaluating} and LiveCodeBench~\citep{jain2024livecodebenchholisticcontaminationfree}.

All downstream evaluations are performed using EvalScope~\citep{evalscope_2024}. We use zero-shot evaluation for all benchmarks except GSM8K, where we adopt a 4-shot setting. For AIME, we report pass@1 and pass@8. During generation, we follow the official inference settings for each reasoning model. For Qwen3-4B-Thinking and Qwen3-8B, we use temperature $0.6$, top-$p$ $0.95$, top-$k$ $20$, min-$p$ $0$, and a maximum model length of $32768$. Qwen3-8B is evaluated with thinking mode enabled.

We compare CodeQuant2 with representative post-training quantization and clustering baselines, including QuaRot~\citep{ashkboos2024quarotoutlierfree4bitinference}, ParoQuant~\citep{liang2025paroquant}, SqueezeLLM~\citep{kim2024squeezellmdenseandsparsequantization}, and CodeQuant~\citep{yin2026codequant}, with BF16 as the full-precision reference. All methods are evaluated under A4W4 and A8W4 settings, where activations use 4 or 8 bits and weights use 4-bit-equivalent capacity, using a 4-bit average assignment-index width across target matrices for CodeQuant2. For fairness, we use the same granularity across methods: per-channel weight quantization and per-token activation quantization. For weight-only baselines such as SqueezeLLM and ParoQuant, we extend them to weight-activation quantization by applying rotation before activation quantization to mitigate activation outliers.

For system evaluation, we measure autoregressive decoding latency on three types of single GPUs: NVIDIA A6000, NVIDIA L40s, and NVIDIA A100. We compare against BF16, QuaRot~\citep{ashkboos2024quarotoutlierfree4bitinference}, and SqueezeLLM~\citep{kim2024squeezellmdenseandsparsequantization} under the A4W4 setting. SqueezeLLM provides a clustered quantization baseline for the latency comparison. Performance is assessed using average prompt and decoding lengths profiled from a subset of examples sampled from WikiText2, GSM8K, and AIME25, with the (prompt, decode) lengths set to (1024, 100), (100, 4000), and (150, 6000), respectively. 

We further specify the method-specific settings used by CodeQuant2. For calibration, we use WikiText2 as the general-text source and generate reasoning-decoding samples from GSM8K training split. We evaluate GSM8K accuracy on the held-out test split. For each GSM8K sample, we first generate 256 tokens using the full-precision model, concatenate the prompt with the generated tokens, and truncate or pad the resulting calibration sequence to 512 tokens. We use a 9:1 mixture of WikiText2 samples and RAC-selected GSM8K samples, ensuring that the calibration set is strictly disjoint from all test sets. The mixed samples are then randomly shuffled to reduce ordering bias. For fine-tuning, we use SGD with momentum $0.9$ and zero weight decay, with separate learning rates of $1\times10^{-4}$ for centroids and $1\times10^{-5}$ for assignment. For adaptive cluster allocation, we use the candidate set $\mathcal{K}=\{8,16,32\}$. All training stages can be conducted on a single NVIDIA H100 GPU. Rotation and scaling fine-tuning requires approximately 30 minutes, adaptive cluster allocation requires around 15 minutes, and the clustering fine-tuning takes about 30 minutes.

\begin{table*}[!t]
\centering
\caption{Mathematical reasoning performance on AIME25 and AIME26 under A4W4 and A8W4, evaluated with both pass@1 and pass@8.}
\label{tab:eval:aime_pass}
\label{tab:eval:aime_a8w4}
\resizebox{\textwidth}{!}{%
\begin{tabular}{ll|cccc|cccc}
\toprule
\multirow{3}{*}{\textbf{Models}} & \multirow{3}{*}{\textbf{Methods}} & \multicolumn{4}{c|}{\textbf{A4W4}} & \multicolumn{4}{c}{\textbf{A8W4}} \\
\cmidrule(lr){3-6} \cmidrule(lr){7-10}
& & \multicolumn{2}{c}{\textbf{AIME25}} & \multicolumn{2}{c|}{\textbf{AIME26}} & \multicolumn{2}{c}{\textbf{AIME25}} & \multicolumn{2}{c}{\textbf{AIME26}} \\
& & \textbf{pass@1} & \textbf{pass@8} & \textbf{pass@1} & \textbf{pass@8} & \textbf{pass@1} & \textbf{pass@8} & \textbf{pass@1} & \textbf{pass@8} \\
\midrule
\multirow{6}{*}{\makecell{Qwen3-4B-\\Thinking}} & BF16 & 0.688 & 0.733 & 0.708 & 0.867 & 0.688 & 0.733 & 0.708 & 0.867 \\
 & QuaRot & 0.004 & 0.033 & 0.000 & 0.000 & 0.175 & 0.233 & 0.088 & 0.133 \\
 & ParoQuant & 0.183 & 0.267 & 0.142 & 0.133 & 0.283 & 0.533 & 0.271 & 0.433 \\
 & SqueezeLLM & 0.192 & 0.333 & 0.158 & 0.200 & 0.438 & 0.600 & 0.454 & 0.600 \\
 & CodeQuant & 0.392 & 0.500 & 0.367 & 0.667 & 0.596 & \textbf{0.833} & 0.638 & 0.800 \\
 & \cellcolor{blue!10}\textbf{Ours} & \cellcolor{blue!10}\textbf{0.563} & \cellcolor{blue!10}\textbf{0.800} & \cellcolor{blue!10}\textbf{0.613} & \cellcolor{blue!10}\textbf{0.800} & \cellcolor{blue!10}\textbf{0.600} & \cellcolor{blue!10}0.800 & \cellcolor{blue!10}\textbf{0.692} & \cellcolor{blue!10}\textbf{0.833} \\
\midrule
\multirow{6}{*}{Qwen3-8B} & BF16 & 0.625 & 0.800 & 0.646 & 0.800 & 0.625 & 0.800 & 0.646 & 0.800 \\
 & QuaRot & 0.033 & 0.100 & 0.004 & 0.033 & 0.200 & 0.333 & 0.133 & 0.267 \\
 & ParoQuant & 0.196 & 0.300 & 0.167 & 0.233 & 0.396 & 0.700 & 0.358 & 0.633 \\
 & SqueezeLLM & 0.242 & 0.367 & 0.179 & 0.300 & 0.404 & 0.567 & 0.446 & 0.700 \\
 & CodeQuant & 0.446 & 0.633 & 0.483 & 0.633 & 0.521 & 0.800 & 0.628 & 0.800 \\
 & \cellcolor{blue!10}\textbf{Ours} & \cellcolor{blue!10}\textbf{0.479} & \cellcolor{blue!10}\textbf{0.700} & \cellcolor{blue!10}\textbf{0.504} & \cellcolor{blue!10}\textbf{0.700} & \cellcolor{blue!10}\textbf{0.613} & \cellcolor{blue!10}\textbf{0.800} & \cellcolor{blue!10}\textbf{0.646} & \cellcolor{blue!10}\textbf{0.833} \\
\midrule
\multirow{6}{*}{\makecell{Phi-4-\\Reasoning}} & BF16 & 0.283 & 0.433 & 0.250 & 0.467 & 0.283 & 0.433 & 0.250 & 0.467 \\
 & QuaRot & 0.192 & 0.267 & 0.129 & 0.300 & 0.221 & 0.367 & 0.213 & 0.367 \\
 & ParoQuant & 0.183 & 0.300 & 0.150 & 0.300 & 0.200 & 0.367 & 0.217 & 0.467 \\
 & SqueezeLLM & \textbf{0.258} & \textbf{0.433} & 0.188 & 0.267 & \textbf{0.288} & 0.400 & 0.208 & 0.300 \\
 & CodeQuant & 0.246 & 0.333 & 0.242 & 0.400 & 0.276 & 0.403 & 0.254 & 0.467 \\
 & \cellcolor{blue!10}\textbf{Ours} & \cellcolor{blue!10}0.246 & \cellcolor{blue!10}0.400 & \cellcolor{blue!10}\textbf{0.250} & \cellcolor{blue!10}\textbf{0.433} & \cellcolor{blue!10}0.279 & \cellcolor{blue!10}\textbf{0.433} & \cellcolor{blue!10}\textbf{0.267} & \cellcolor{blue!10}\textbf{0.467} \\
\bottomrule
\end{tabular}%
}
\par\vspace{-10pt}
\end{table*}

\subsection{Accuracy Evaluation}
Table~\ref{tab:eval:main} reports the A4W4 results across Qwen3-4B-Thinking, Qwen3-8B, and Phi-4-Reasoning. Overall, CodeQuant2 achieves the strongest performance, outperforming QuaRot, SqueezeLLM, ParoQuant, and CodeQuant on most models and tasks. The improvement is most significant on Qwen3-4B-Thinking, where CodeQuant2 increases average accuracy by 10.6\% points over CodeQuant, suggesting that it is especially effective under aggressive quantization where both activation outliers and clustering errors become severe. On Qwen3-8B and Phi-4-Reasoning, the gains are more moderate but remain consistent, indicating that CodeQuant2 generalizes across different reasoning models. Importantly, the improvements on reasoning and code-generation benchmarks do not sacrifice general language modeling quality, as CodeQuant2 maintains competitive perplexity on WikiText2 and C4. The A8W4 results are provided in Table~\ref{tab:eval:acc_a8w4} in the appendix.

We further test quantized reasoning models on challenging mathematical benchmarks, including AIME25~\citep{aime25} and AIME26~\citep{aime26}. These tasks are highly sensitive to quantization because they require long multi-step reasoning and accurate handling of numerical and symbolic information. Besides pass@1, we also report pass@8, where pass@$k$ indicates whether at least one of $k$ sampled responses gives the correct answer.
As shown in Table~\ref{tab:eval:aime_pass}, CodeQuant2 consistently improves mathematical reasoning robustness under the challenging A4W4 setting. The gains are especially clear on Qwen3 reasoning models, where existing quantized baselines suffer large drops on AIME problems. CodeQuant2 also improves pass@8, so at least one of eight sampled responses is more likely to be correct. On Phi-4-Reasoning, CodeQuant2 also improves over CodeQuant on AIME25 pass@8 (0.400 vs. 0.333) and AIME26 pass@8 (0.433 vs. 0.400). The corresponding A8W4 results appear alongside the A4W4 results in Table~\ref{tab:eval:aime_a8w4}.

We further evaluate code generation on HumanEval and LiveCodeBench, which test not only reasoning ability but also syntactic correctness and instruction following, making them a useful measure of whether low-bit quantization preserves broader generation capabilities. Under A4W4, CodeQuant2 maintains strong code performance while improving reasoning-heavy tasks. The largest gain appears on LiveCodeBench for Qwen3-4B-Thinking, where CodeQuant2 outperforms CodeQuant by 15.0\% points. These results show gains on coding as well as math benchmarks under A4W4.

Although CodeQuant2 targets reasoning-oriented quantization, it preserves strong general-task performance. On language understanding and commonsense reasoning benchmarks such as ARC, PIQA, and WinoGrande, CodeQuant2 performs comparably to or better than existing quantized baselines. Detailed results are provided in Table~\ref{tab:eval:general_task} in the appendix.


\begin{table*}[t]\centering
\begin{minipage}[t]{0.36\textwidth}
\centering
\captionsetup{justification=raggedright,singlelinecheck=false}
\caption{Qwen3-4B-Thinking on L40s: decode throughput.}
\label{tab:eval:throughput_different_cluster_number}
\resizebox{\linewidth}{!}{
\begin{tabular}{c|c|c}
\toprule
\textbf{Cluster Num} & \textbf{GSM8K} & \textbf{AIME25} \\
\midrule
8 & 47 token/s & 36 token/s \\
16 & 42 token/s & 30 token/s \\
32 & 33 token/s & 25 token/s \\
\cellcolor{blue!10}\textbf{Ours}
& \cellcolor{blue!10}\textbf{44 token/s}
& \cellcolor{blue!10}\textbf{32 token/s} \\
\bottomrule
\end{tabular}
}
\par\vspace{-10pt}
\medskip
\captionof{table}{Qwen3-4B-Thinking on A6000: decode speedup over BF16.}
\label{tab:eval:batch_size_speedup}
{\footnotesize
\setlength{\tabcolsep}{2pt}
\begin{adjustbox}{max width=\linewidth}
\begin{tabular}{c|ccc}
\toprule
\textbf{Batch} & \textbf{WikiText2} & \textbf{GSM8K} & \textbf{AIME25} \\
\midrule
1 & 2.31 & 2.45 & 2.48 \\
4 & 2.32 & 2.46 & 2.48 \\
8 & 2.30 & 2.44 & 2.46 \\
16 & 2.28 & 2.43 & 2.45 \\
\bottomrule
\end{tabular}
\end{adjustbox}
}
\end{minipage}\hfill
\begin{minipage}[t]{0.63\textwidth}
\centering
\vspace{0pt}
\includegraphics[width=\linewidth]{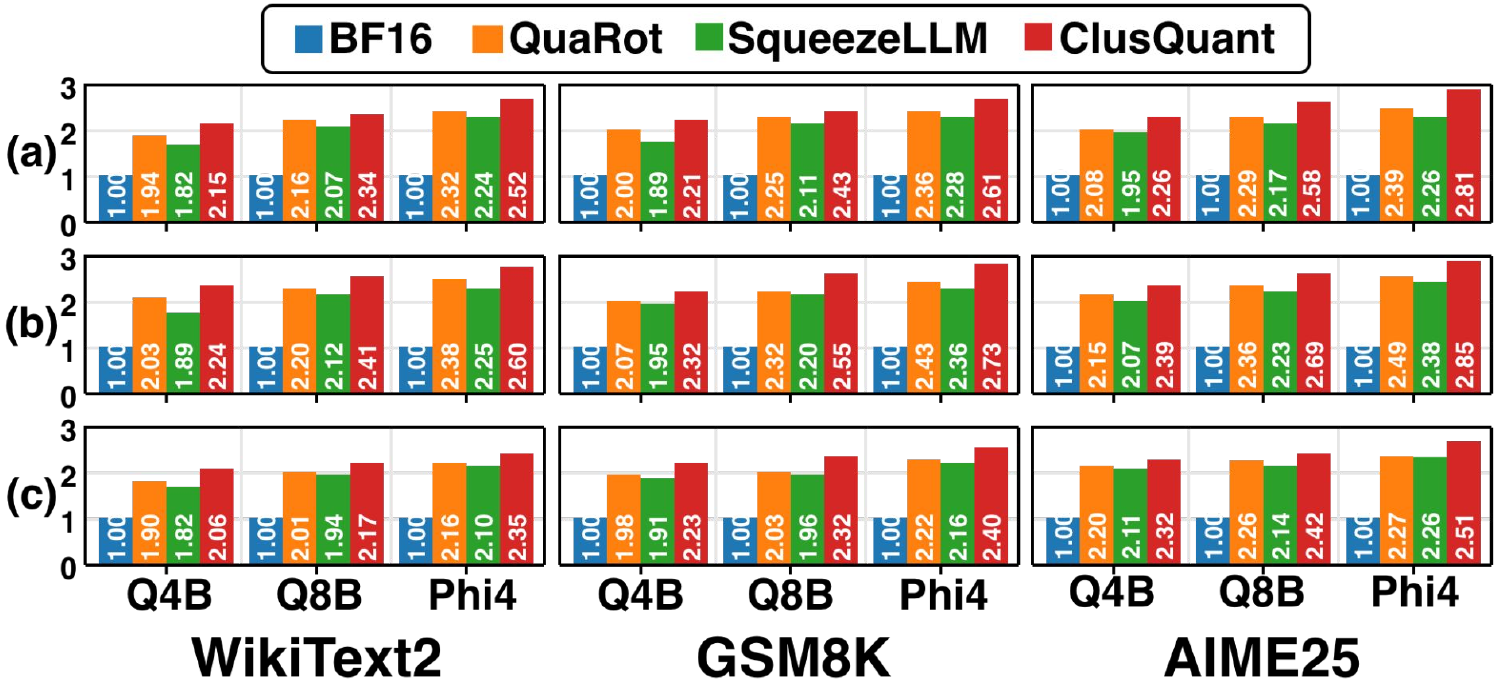}
\captionof{figure}{Normalized decoding speedup on (a) A6000, (b) L40s, and (c) A100 GPU across WikiText2-, GSM8K-, and AIME25-based workloads. Q4B, Q8B, and Phi4 denote Qwen3-4B-Thinking, Qwen3-8B, and Phi4-Reasoning, respectively.}
\label{fig:lut_hw_eval}
\end{minipage}
\par\vspace{-10pt}
\end{table*}

\subsection{System Evaluation}
Figure~\ref{fig:lut_hw_eval} presents the normalized speedups of all baselines, with BF16 latency normalized to 1. Compared with the BF16 models, CodeQuant2 achieves an average $2.42\times$ speedup, demonstrating the effectiveness of low-bit activation and weight quantization together with the LUT-based GEMM design. The improvement over SqueezeLLM further indicates that clustering-based quantization can benefit from a dedicated GPU implementation with optimized LUT operations. Considering the accuracy results of CodeQuant2 shown in Table~\ref{tab:eval:main}, CodeQuant2 achieves a strong accuracy--efficiency trade-off among the evaluated baselines. Table~\ref{tab:eval:batch_size_speedup} shows that CodeQuant2 retains at least a $2.28\times$ speedup on A6000 at batch size 16, indicating that the
gain persists under batching.

\subsection{Ablation Study}

\paragraph{Component-wise Ablation}

\begin{table}[h]
\centering
\caption{Cumulative component ablation on Qwen3-4B-Thinking under A4W4. The base configuration uses fixed random rotation and SmoothQuant scaling. Each subsequent row adds the listed component cumulatively. AIME results are pass@1, and gains are in percentage points.}
\label{tab:ablation:components}
\resizebox{\linewidth}{!}{
\begin{tabular}{l|cccccc|cc}
\toprule
\textbf{Configuration}
& \textbf{MATH-500}
& \textbf{GSM8K}
& \textbf{AIME25}
& \textbf{AIME26}
& \textbf{HumanEval}
& \textbf{LiveCodeBench}
& \textbf{Avg. Acc.}
& \textbf{Gain} \\
\midrule
Base   & 0.930 & 0.929 & 0.392 & 0.367 & 0.902 & 0.464 & 0.664 & -- \\
+ R\&S & 0.928 & 0.934 & 0.450 & 0.421 & 0.933 & 0.557 & 0.704 & +4.00 \\
+ ACA  & 0.938 & 0.938 & 0.542 & 0.488 & 0.937 & 0.599 & 0.740 & +3.64 \\
+ LC   & 0.950 & 0.937 & 0.546 & 0.567 & 0.939 & 0.619 & 0.760 & +1.94 \\
\rowcolor{blue!10} + RAC  & 0.950 & 0.938 & 0.563 & 0.613 & 0.945 & 0.614 & \textbf{0.770} & +1.08 \\
\bottomrule
\end{tabular}
}
\par\vspace{-5pt}
\end{table}

To evaluate the contribution of each technique in Section~\ref{sec:methdology}, Table~\ref{tab:ablation:components} reports the cumulative ablation results on Qwen3-4B-Thinking under A4W4. Without the four techniques, the average reasoning accuracy is 0.664. Fine-tuned rotation and scaling (R\&S) increases it to 0.704, highlighting the importance of outlier mitigation for stable low-bit reasoning quantization. Adding adaptive cluster allocation (ACA) improves accuracy to 0.740, a gain of 3.64\% points, consistent with the benefit of varying cluster capacity across matrices. Learnable clustering (LC) further increases accuracy to 0.760, showing the benefit of optimizing cluster assignments. Finally, reasoning-aware calibration (RAC) raises accuracy to 0.770, confirming that reasoning-time decoded tokens provide useful calibration signals beyond standard language-modeling data.

\paragraph{Effect of Calibration Selection}\label{abl:calibration}
\begin{table*}[!ht]
\centering
\caption{Effect of decoding calibration under different calibration selections. M500, GSM, A25, A26, HE, and LCB denote MATH-500, GSM8K, AIME25, AIME26, HumanEval, and LiveCodeBench. In panel (a), A25 and A26 are pass@1, while A25@8 and A26@8 are pass@8.}
\label{tab:ablation:decoding_calibration}
{\footnotesize
\begin{minipage}[t]{0.57\textwidth}
\centering
\textbf{(a) Segment selection}\par\smallskip
\setlength{\tabcolsep}{2pt}
\begin{adjustbox}{max width=\linewidth}
\begin{tabular}{l|ccccccccc}
\toprule
\textbf{Segment} & \textbf{M500} & \textbf{GSM} & \textbf{A25} & \textbf{A25@8}& \textbf{A26} & \textbf{A26@8} & \textbf{HE} & \textbf{LCB} & \textbf{Avg.} \\
\midrule
None & 0.950 & 0.937 & 0.546 & 0.667 & 0.567 & 0.733 & 0.939 & 0.619 & 0.745 \\
\rowcolor{blue!10} First & 0.942 & 0.934 & 0.517 & 0.733 & 0.600 & 0.767 & 0.927 & 0.609 & \textbf{0.754} \\
Middle & 0.942 & 0.942 & 0.517 & 0.700 & 0.529 & 0.733 & 0.922 & 0.578 & 0.733 \\
Last & 0.936 & 0.936 & 0.517 & 0.667 & 0.554 & 0.700 & 0.933 & 0.590 & 0.729 \\
All 768 & 0.944 & 0.941 & 0.467 & 0.667 & 0.496 & 0.667 & 0.921 & 0.560 & 0.708 \\
\bottomrule
\end{tabular}
\end{adjustbox}
\end{minipage}\hfill
\begin{minipage}[t]{0.42\textwidth}
\centering
\textbf{(b) Prefix length}\par\smallskip
\setlength{\tabcolsep}{2pt}
\begin{adjustbox}{max width=\linewidth}
\begin{tabular}{c|ccccccc}
\toprule
\textbf{Tokens} & \textbf{M500} & \textbf{GSM} & \textbf{A25} & \textbf{A26} & \textbf{HE} & \textbf{LCB} & \textbf{Avg.} \\
\midrule
64 & 0.912 & 0.923 & 0.358 & 0.354 & 0.884 & 0.392 & 0.637 \\
128 & 0.936 & 0.939 & 0.521 & 0.567 & 0.933 & 0.549 & 0.741 \\
\rowcolor{blue!10} 256 & 0.950 & 0.938 & 0.563 & 0.613 & 0.945 & 0.614 & \textbf{0.770} \\
512 & 0.938 & 0.938 & 0.529 & 0.571 & 0.915 & 0.591 & 0.747 \\
768 & 0.944 & 0.941 & 0.467 & 0.496 & 0.921 & 0.560 & 0.721 \\
\bottomrule
\end{tabular}
\end{adjustbox}
\end{minipage}
}
\par\vspace{-5pt}
\end{table*}

Table~\ref{tab:ablation:decoding_calibration} tests two design choices for decoding calibration on Qwen3-4B-Thinking under A4W4: which 256-token segment of a reasoning trace to calibrate on, and how many initial decoded tokens to include. For segment selection in panel (a), we generate 768 tokens from GSM8K prompts with the full-precision model, and compute loss only on tokens 1--256, 257--512, or 513--768 while masking the other tokens. We also evaluate no decoding calibration and all 768 tokens. For prefix length in panel (b), we mix GSM8K decoding samples with WikiText2 calibration data, as in the component ablation in Table~\ref{tab:ablation:components}. The loss includes the WikiText2 samples and a prefix of 64, 128, 256, 512, or 768 decoded tokens from each GSM8K sample. The 256-token setting matches the calibration of our reported model and reproduces its task scores. We believe that including later tokens may dilute the influence of high-variance early activations on calibration, making it less effective at mitigating their outliers and leading to the accuracy drop at 768 tokens.

For segment selection, the first 256 tokens yield the highest eight-score average among the three segments at 0.754, versus 0.733 for the middle and 0.729 for the last. They also exceed the no-decoding baseline of 0.745. The gain over no decoding is concentrated in AIME pass@8: AIME25 rises from 0.667 to 0.733, and AIME26 rises from 0.733 to 0.767. AIME25 pass@1 instead falls from 0.546 to 0.517, so the improvement does not hold for every metric. This pattern is consistent with the stronger early activation fluctuations in Figure~\ref{fig:calibration_motivation}. Using all 768 tokens lowers the eight-score average to 0.708, below every individual segment.

For prefix length, with WikiText2 included in calibration, the six-task average rises from 0.637 at 64 decoded tokens to 0.770 at 256, then falls to 0.747 at 512 and 0.721 at 768. From 256 to 768 tokens, AIME25 drops from 0.563 to 0.467 and AIME26 from 0.613 to 0.496, while GSM8K stays near 0.94. 

\paragraph{Comparison of Cluster Allocation Strategies}
Table~\ref{tab:eval:throughput_different_cluster_number} reports decoding throughput under different cluster allocation strategies. We compare fixed cluster counts of 8, 16, and 32 with our adaptive cluster allocation (ACA) method from Section~\ref{sec:aca}, whose average assignment-index width across target matrices is 4 bits, as in the fixed 16-cluster setting. The fixed settings reveal a clear efficiency trade-off: throughput drops from 47 tokens/s with 8 clusters to 42 tokens/s with 16 clusters and 33 tokens/s with 32 clusters on the GSM8K dataset. ACA achieves 44 tokens/s, outperforming the fixed 16-cluster setting and approaching the efficiency of the fixed 8-cluster setting. ACA is also faster on AIME25. Its accuracy gain in Table~\ref{tab:ablation:components} and its higher throughput are consistent with reallocating clusters toward sensitive matrices under a profiled-latency budget.

\section{Conclusion and Limitations}
In conclusion, CodeQuant2 combines early decoded tokens with general samples to build a reasoning-aware calibration set that captures the activation fluctuations at the start of reasoning. Using this calibration set, CodeQuant2 learns rotation and scaling transformations that smooth activation outliers. For the weights, it allocates cluster budgets across weight matrices to balance accuracy and latency, and then fine-tunes the centroid values and assignments to better match the full-precision outputs. Finally, a LUT-based CUDA kernel operates directly on the clustered representation during inference.

Although CodeQuant2 performs well on the evaluated reasoning tasks, further study of its generated reasoning trajectories could provide a more complete picture of the model's reasoning behavior after quantization. Such an analysis could examine whether a correct answer is accompanied by repetition or unnecessarily long reasoning. It could also explore how quantization interacts with behaviors learned during post-training beyond those measured by our benchmarks, as well as potential safety implications such as unreliable or overconfident outputs. We have not evaluated these questions in this work and believe addressing them is a promising direction for future work.

\section*{Funding Statement}
The authors declare no funding received for this paper.

\section*{Competing Interests Statement}
The authors declare no competing interests.

\bibliographystyle{plain}
\bibliography{ref}

\appendix

\section{Additional Activation Analysis}
\label{sec:app:activation_fluctuation}
Table~\ref{tab:app:activation_fluctuation} extends Figure~\ref{fig:calibration_motivation} to two other reasoning models and two other math datasets. The model rows use GSM8K prompts, and the dataset rows use Qwen3-4B-Thinking. Step 0 in this table is the first decoded token, labeled step 1 in Figure~\ref{fig:calibration_motivation}. We take the point where each activation-standard-deviation curve begins to level off (the elbow) as its stabilization step. The final column reports the percentage increase from the curve's value at the elbow to its peak.

\begin{table}[H]
\centering
\caption{Early-stage activation fluctuation across additional reasoning models and datasets.}
\label{tab:app:activation_fluctuation}
{\small
\setlength{\tabcolsep}{3pt}
\begin{adjustbox}{max width=\columnwidth}
\begin{tabular}{llccc}
\toprule
\textbf{Comparison} & \textbf{Setting} & \textbf{Peak step} & \textbf{Elbow step} & \textbf{Peak above elbow} \\
\midrule
Model & GPT-OSS-20B & 0 & 23 & +24.1\% \\
Model & Qwen3-30B-A3B-Thinking & 0 & 34 & +38.7\% \\
Dataset & MATH-500 & 0 & 26 & +17.1\% \\
Dataset & AIME24 & 5 & 39 & +20.5\% \\
\bottomrule
\end{tabular}
\end{adjustbox}
}
\end{table}

\section{Adaptive Cluster Importance Score}
\label{sec:app:importance_score}
In adaptive cluster allocation, we estimate the importance of each candidate cluster number change by measuring how much its induced output perturbation affects the local reconstruction loss.
Let the local reconstruction loss be
\[
\mathcal{L}_{\mathrm{recon}}(\widetilde{\mathbf{Y}})
=
\left\|
\mathbf{Y}_{\mathrm{fp}}-\widetilde{\mathbf{Y}}
\right\|^2 .
\]
When the cluster number is changed from $K_{\mathrm{ori}}$ to $K_{\mathrm{tar}}$, the clustered output changes from $\widetilde{\mathbf{Y}}^{(K_{\mathrm{ori}})}$ to $\widetilde{\mathbf{Y}}^{(K_{\mathrm{tar}})}$, producing the output perturbation
\[
\Delta \mathbf{Y}_{K_{\mathrm{ori}}\rightarrow K_{\mathrm{tar}}}
=
\widetilde{\mathbf{Y}}^{(K_{\mathrm{ori}})}
-
\widetilde{\mathbf{Y}}^{(K_{\mathrm{tar}})} .
\]
Using a first-order Taylor approximation around
$\widetilde{\mathbf{Y}}^{(K_{\mathrm{ori}})}$, the change in reconstruction loss can be estimated as
\[
\Delta \mathcal{L}_{\mathrm{recon}}
\approx -
\left\langle
\nabla_{\widetilde{\mathbf{Y}}}\mathcal{L}_{\mathrm{recon}},
\Delta \mathbf{Y}_{K_{\mathrm{ori}}\rightarrow K_{\mathrm{tar}}}
\right\rangle .
\]
where $\langle \cdot,\cdot\rangle$ denotes the inner product.
To obtain a stable matrix-wise score, we use the squared element-wise contribution and normalize it by the number of calibration tokens $N$:
\[
I_{K_{\mathrm{ori}}\rightarrow K_{\mathrm{tar}}}
=
\frac{1}{N}
\left\|
\Delta\mathbf{Y}_{K_{\mathrm{ori}}\rightarrow K_{\mathrm{tar}}}
\odot
\nabla_{\widetilde{\mathbf{Y}}}
\mathcal{L}_{\mathrm{recon}}
\right\|_F^2.
\]
This score measures how strongly the output perturbation caused by changing the cluster number affects the local reconstruction loss.

\section{Additional Evaluation Results}

\subsection{General Tasks}
Table~\ref{tab:eval:general_task} reports the results on standard language understanding and commonsense reasoning tasks. Overall, our method preserves strong general-task performance under both A4W4 and A8W4 settings. Under A4W4, our method has the highest average score among quantized methods on all three models. Compared with CodeQuant, our method improves the average score from 0.845 to 0.856 on Qwen3-4B-Thinking and from 0.854 to 0.856 on Qwen3-8B. On Phi-4-Reasoning, the average rises from 0.765 to 0.766.
Under the A8W4 setting, our method further reduces the gap to the BF16 reference and achieves the best average score among quantized methods across all three models.

These results indicate that our method improves reasoning-oriented quantization performance while maintaining competitive performance on general language understanding and commonsense reasoning tasks. The evaluated general tasks show that the gains extend beyond reasoning-heavy benchmarks.

\begin{table}[!t]
\centering
\caption{Performance on standard language understanding and commonsense reasoning tasks under A4W4 and A8W4 quantization.}
\resizebox{0.99\columnwidth}{!}{
\begin{tabular}{c|c|cccc}
\toprule
\textbf{Models} 
& \textbf{Methods} 
& \textbf{ARC} ($\uparrow$) 
& \textbf{PIQA} ($\uparrow$) 
& \textbf{WinoGrande} ($\uparrow$) 
& \textbf{Avg} ($\uparrow$) \\
\midrule

\multicolumn{6}{c}{\textbf{A4W4}} \\
\midrule

\multirow{6}{*}{\makecell{Qwen3-\\4B-Thinking}}
& BF16       & 0.937 & 0.874 & 0.830 & 0.881 \\
& QuaRot     & 0.752 & 0.562 & 0.281 & 0.532 \\
& ParoQuant  & 0.913 & 0.789 & 0.641 & 0.781 \\
& SqueezeLLM & 0.914 & 0.791 & 0.681 & 0.795 \\
& CodeQuant  & 0.928 & 0.836 & 0.772 & 0.845 \\
& \cellcolor{blue!10}\textbf{Ours}
             & \cellcolor{blue!10}\textbf{0.932}
             & \cellcolor{blue!10}\textbf{0.851}
             & \cellcolor{blue!10}\textbf{0.785}
             & \cellcolor{blue!10}\textbf{0.856} \\

\cmidrule(lr){1-6}

\multirow{6}{*}{\makecell{Qwen3-\\8B}}
& BF16       & 0.942 & 0.904 & 0.818 & 0.888 \\
& QuaRot     & 0.822 & 0.631 & 0.358 & 0.604 \\
& ParoQuant  & 0.847 & 0.475 & 0.520 & 0.614 \\
& SqueezeLLM & 0.921 & 0.823 & 0.677 & 0.807 \\
& CodeQuant  & \textbf{0.936} & 0.871 & 0.755 & 0.854 \\
& \cellcolor{blue!10}\textbf{Ours}
             & \cellcolor{blue!10}0.930
             & \cellcolor{blue!10}\textbf{0.872}
             & \cellcolor{blue!10}\textbf{0.766}
             & \cellcolor{blue!10}\textbf{0.856} \\

\cmidrule(lr){1-6}

\multirow{6}{*}{\makecell{Phi-4-\\Reasoning}}
& BF16       & 0.837 & 0.868 & 0.780 & 0.828 \\
& QuaRot     & 0.758 & 0.745 & 0.659 & 0.721 \\
& ParoQuant  & \textbf{0.767} & 0.758 & 0.676 & 0.733 \\
& SqueezeLLM & 0.740 & 0.760 & 0.685 & 0.728 \\
& CodeQuant  & 0.745 & \textbf{0.821} & 0.729 & 0.765 \\
& \cellcolor{blue!10}\textbf{Ours}
             & \cellcolor{blue!10}0.756
             & \cellcolor{blue!10}0.811
             & \cellcolor{blue!10}\textbf{0.731}
             & \cellcolor{blue!10}\textbf{0.766} \\

\midrule

\multicolumn{6}{c}{\textbf{A8W4}} \\
\midrule

\multirow{6}{*}{\makecell{Qwen3-\\4B-Thinking}}
& BF16       & 0.937 & 0.874 & 0.830 & 0.881 \\
& QuaRot     & 0.913 & 0.798 & 0.582 & 0.764 \\
& ParoQuant  & 0.930 & 0.840 & 0.744 & 0.838 \\
& SqueezeLLM & 0.936 & 0.846 & 0.785 & 0.855 \\
& CodeQuant  & 0.936 & 0.856 & \textbf{0.815} & 0.869 \\
& \cellcolor{blue!10}\textbf{Ours}
             & \cellcolor{blue!10}\textbf{0.937}
             & \cellcolor{blue!10}\textbf{0.864}
             & \cellcolor{blue!10}0.807
             & \cellcolor{blue!10}\textbf{0.870} \\

\cmidrule(lr){1-6}

\multirow{6}{*}{\makecell{Qwen3-\\8B}}
& BF16       & 0.942 & 0.904 & 0.818 & 0.888 \\
& QuaRot     & 0.915 & 0.862 & 0.642 & 0.806 \\
& ParoQuant  & 0.920 & 0.845 & 0.721 & 0.829 \\
& SqueezeLLM & 0.938 & 0.874 & 0.777 & 0.863 \\
& CodeQuant  & 0.938 & 0.890 & 0.796 & 0.875 \\
& \cellcolor{blue!10}\textbf{Ours}
             & \cellcolor{blue!10}\textbf{0.939}
             & \cellcolor{blue!10}\textbf{0.890}
             & \cellcolor{blue!10}\textbf{0.798}
             & \cellcolor{blue!10}\textbf{0.876} \\

\cmidrule(lr){1-6}

\multirow{6}{*}{\makecell{Phi-4-\\Reasoning}}
& BF16       & 0.837 & 0.868 & 0.780 & 0.828 \\
& QuaRot     & 0.783 & 0.774 & 0.709 & 0.755 \\
& ParoQuant  & 0.786 & 0.780 & 0.702 & 0.756 \\
& SqueezeLLM & 0.790 & 0.836 & 0.757 & 0.794 \\
& CodeQuant  & 0.782 & 0.816 & 0.731 & 0.776 \\
& \cellcolor{blue!10}\textbf{Ours}
             & \cellcolor{blue!10}\textbf{0.847}
             & \cellcolor{blue!10}\textbf{0.867}
             & \cellcolor{blue!10}\textbf{0.785}
             & \cellcolor{blue!10}\textbf{0.833} \\

\bottomrule
\end{tabular}
}
\label{tab:eval:general_task}
\end{table}

\subsection{A8W4 Results}

\paragraph{Main Results.}
Table~\ref{tab:eval:acc_a8w4} reports the results under the A8W4 setting. Compared with A4W4, A8W4 provides higher activation precision and therefore preserves stronger model performance across most tasks. Overall, our method achieves the best average accuracy among all quantized methods on all three reasoning models. Specifically, our method obtains average scores of 0.808, 0.790, and 0.593 on Qwen3-4B-Thinking, Qwen3-8B, and Phi-4-Reasoning, respectively.

On Qwen3-4B-Thinking, our method improves the average score from 0.779 to 0.808 compared with CodeQuant, while also achieving the best perplexity on both WikiText2 and C4 among quantized methods. It also provides clear improvements on reasoning-heavy benchmarks, especially AIME26 and LiveCodeBench. On Qwen3-8B, our method obtains the highest average score of 0.790 and outperforms CodeQuant on GSM8K, AIME25, AIME26, and LiveCodeBench, although CodeQuant achieves slightly lower perplexity. On Phi-4-Reasoning, our method achieves the best average score and the best results on WikiText2, C4, MATH500, GSM8K, AIME26, and HumanEval among the quantized methods.

These results show that our method remains effective under the less aggressive A8W4 setting. In particular, it improves reasoning and code generation performance while maintaining competitive language modeling quality.

\begin{table}[!t]
\caption{Main evaluation results under A8W4 quantization.}
\centering
\resizebox{1\columnwidth}{!}{
\begin{tabular}{c|c|cc|cccccc|c}
    \toprule
    
    \textbf{Models} & \textbf{Methods} 
    & \textbf{Wiki2} ($\downarrow$) 
    & \textbf{C4} ($\downarrow$) 
    & \textbf{MA500} ($\uparrow$) 
    & \textbf{GSM8K} ($\uparrow$) 
    & \textbf{AIME25} ($\uparrow$) 
    & \textbf{AIME26} ($\uparrow$) 
    & \textbf{HE} ($\uparrow$) 
    & \textbf{LCB} ($\uparrow$) 
    & \textbf{Avg} ($\uparrow$) \\
    
    \midrule

    \multicolumn{11}{c}{\textbf{A8W4}} \\
    \midrule
    
    \multirow{6}{*}{\makecell{Qwen3-\\4B-Thinking}} 
    & BF16       & 11.478 & 18.391 & 0.958 & 0.952 & 0.688 & 0.708 & 0.963 & 0.758 & 0.838 \\
    & QuaRot     & 14.312 & 22.492 & 0.758 & 0.868 & 0.175 & 0.088 & 0.537 & 0.137 & 0.427 \\
    & ParoQuant  & 13.444 & 21.190 & 0.900 & 0.920 & 0.283 & 0.271 & 0.811 & 0.448 & 0.606 \\
    & SqueezeLLM & 12.999 & 20.260 & 0.940 & 0.938 & 0.438 & 0.454 & 0.915 & 0.552 & 0.706 \\
    & CodeQuant  & 12.430 & 19.732 & 0.946 & 0.951 & 0.596 & 0.638 & 0.915 & 0.629 & 0.779 \\
    & \cellcolor{blue!10}\textbf{Ours}
                 & \cellcolor{blue!10}\textbf{12.081}
                 & \cellcolor{blue!10}\textbf{19.206}
                 & \cellcolor{blue!10}\textbf{0.960}
                 & \cellcolor{blue!10}\textbf{0.951}
                 & \cellcolor{blue!10}\textbf{0.600}
                 & \cellcolor{blue!10}\textbf{0.692}
                 & \cellcolor{blue!10}\textbf{0.957}
                 & \cellcolor{blue!10}\textbf{0.686}
                 & \cellcolor{blue!10}\textbf{0.808} \\
    
    \cmidrule(lr){1-11}
    
    \multirow{6}{*}{\makecell{Qwen3-\\8B}} 
    & BF16       & 10.047 & 15.439 & 0.948 & 0.958 & 0.625 & 0.646 & 0.921 & 0.732 & 0.805 \\
    & QuaRot     & 12.471 & 18.507 & 0.784 & 0.880 & 0.200 & 0.133 & 0.256 & 0.019 & 0.379 \\
    & ParoQuant  & 11.337 & 17.234 & 0.908 & 0.933 & 0.396 & 0.358 & 0.881 & 0.463 & 0.656 \\
    & SqueezeLLM & 10.798 & 16.580 & 0.926 & 0.941 & 0.404 & 0.446 & \textbf{0.933} & 0.544 & 0.699 \\
    & CodeQuant  & \textbf{10.216} & 16.883 & \textbf{0.942} & 0.947 & 0.521 & 0.628 & 0.921 & 0.648 & 0.768 \\
    & \cellcolor{blue!10}\textbf{Ours}
                 & \cellcolor{blue!10}11.033
                 & \cellcolor{blue!10}\textbf{16.490}
                 & \cellcolor{blue!10}0.938
                 & \cellcolor{blue!10}\textbf{0.955}
                 & \cellcolor{blue!10}\textbf{0.613}
                 & \cellcolor{blue!10}\textbf{0.646}
                 & \cellcolor{blue!10}0.915
                 & \cellcolor{blue!10}\textbf{0.671}
                 & \cellcolor{blue!10}\textbf{0.790} \\

    \cmidrule(lr){1-11}
    
    \multirow{6}{*}{\makecell{Phi-4-\\Reasoning}} 
    & BF16       & 7.393 & 13.187 & 0.858 & 0.943 & 0.283 & 0.250 & 0.872 & 0.410 & 0.603 \\
    & QuaRot     & 8.512 & 14.687 & 0.808 & 0.937 & 0.221 & 0.213 & 0.652 & 0.290 & 0.520 \\
    & ParoQuant  & 7.801 & 13.865 & 0.832 & 0.927 & 0.200 & 0.217 & 0.707 & 0.303 & 0.531 \\
    & SqueezeLLM & 7.759 & 13.763 & 0.838 & 0.942 & \textbf{0.288} & 0.208 & 0.835 & 0.373 & 0.581 \\
    & CodeQuant  & 7.755 & 13.768 & 0.822 & 0.942 & 0.276 & 0.254 & 0.823 & \textbf{0.384} & 0.584 \\
    & \cellcolor{blue!10}\textbf{Ours}
                 & \cellcolor{blue!10}\textbf{7.699}
                 & \cellcolor{blue!10}\textbf{13.592}
                 & \cellcolor{blue!10}\textbf{0.844}
                 & \cellcolor{blue!10}\textbf{0.944}
                 & \cellcolor{blue!10}0.279
                 & \cellcolor{blue!10}\textbf{0.267}
                 & \cellcolor{blue!10}\textbf{0.848}
                 & \cellcolor{blue!10}0.375
                 & \cellcolor{blue!10}\textbf{0.593} \\
    
    \bottomrule
            
\end{tabular}
}
\label{tab:eval:acc_a8w4}
\end{table}

\paragraph{AIME Results.}
The A8W4 columns of Table~\ref{tab:eval:aime_a8w4} report pass@1 and pass@8 results on the AIME mathematical reasoning benchmarks. Overall, our method achieves strong performance across both single-sample and multi-sample evaluation. On Qwen3-4B-Thinking, our method obtains the best pass@1 results on both AIME25 and AIME26, improving AIME26 pass@1 from 0.638 to 0.692 compared with CodeQuant. On Qwen3-8B, our method achieves the best or tied-best results on all four metrics. In particular, it improves AIME25 pass@1 from 0.521 to 0.613 and AIME26 pass@8 from 0.800 to 0.833 compared with CodeQuant. For Phi-4-Reasoning, the performance differences among strong baselines are smaller, while our method still achieves the best AIME26 pass@1 result and matches the best AIME25 pass@8 and AIME26 pass@8 results. These results show that our method better preserves mathematical reasoning ability under A8W4 quantization, especially on pass@1 metrics that reflect the quality of a single generated solution.

\newpage
\section*{NeurIPS Paper Checklist}

The checklist is designed to encourage best practices for responsible machine learning research, addressing issues of reproducibility, transparency, research ethics, and societal impact. Do not remove the checklist: {\bf The papers not including the checklist will be desk rejected.} The checklist should follow the references and follow the (optional) supplemental material.  The checklist does NOT count towards the page
limit. 

Please read the checklist guidelines carefully for information on how to answer these questions. For each question in the checklist:
\begin{itemize}
    \item You should answer \answerYes{}, \answerNo{}, or \answerNA{}.
    \item \answerNA{} means either that the question is Not Applicable for that particular paper or the relevant information is Not Available.
    \item Please provide a short (1--2 sentence) justification right after your answer (even for \answerNA). 
\end{itemize}

{\bf The checklist answers are an integral part of your paper submission.} They are visible to the reviewers, area chairs, senior area chairs, and ethics reviewers. You will also be asked to include it (after eventual revisions) with the final version of your paper, and its final version will be published with the paper.

The reviewers of your paper will be asked to use the checklist as one of the factors in their evaluation. While \answerYes{} is generally preferable to \answerNo{}, it is perfectly acceptable to answer \answerNo{} provided a proper justification is given (e.g., error bars are not reported because it would be too computationally expensive'' or ``we were unable to find the license for the dataset we used''). In general, answering \answerNo{} or \answerNA{} is not grounds for rejection. While the questions are phrased in a binary way, we acknowledge that the true answer is often more nuanced, so please just use your best judgment and write a justification to elaborate. All supporting evidence can appear either in the main paper or the supplemental material, provided in appendix. If you answer \answerYes{} to a question, in the justification please point to the section(s) where related material for the question can be found.

IMPORTANT, please:
\begin{itemize}
    \item {\bf Delete this instruction block, but keep the section heading ``NeurIPS Paper Checklist"},
    \item  {\bf Keep the checklist subsection headings, questions/answers and guidelines below.}
    \item {\bf Do not modify the questions and only use the provided macros for your answers}.
\end{itemize}


\begin{enumerate}

\item {\bf Claims}
    \item[] Question: Do the main claims made in the abstract and introduction accurately reflect the paper's contributions and scope?
    \item[] Answer: \answerYes{} 
    \item[] Justification: The abstract and introduction clearly state the main contributions of CodeQuant2, including reasoning-aware calibration, clustering-based low-bit quantization, adaptive cluster allocation, and LUT-based CUDA execution. The claimed accuracy and efficiency improvements are supported by the experimental results and system evaluation in the paper.
    \item[] Guidelines:
    \begin{itemize}
        \item The answer \answerNA{} means that the abstract and introduction do not include the claims made in the paper.
        \item The abstract and/or introduction should clearly state the claims made, including the contributions made in the paper and important assumptions and limitations. A \answerNo{} or \answerNA{} answer to this question will not be perceived well by the reviewers. 
        \item The claims made should match theoretical and experimental results, and reflect how much the results can be expected to generalize to other settings. 
        \item It is fine to include aspirational goals as motivation as long as it is clear that these goals are not attained by the paper. 
    \end{itemize}

\item {\bf Limitations}
    \item[] Question: Does the paper discuss the limitations of the work performed by the authors?
    \item[] Answer: \answerYes{} 
    \item[] Justification: We discuss the limitation and risk in the last section of the paper.
    \item[] Guidelines:
    \begin{itemize}
        \item The answer \answerNA{} means that the paper has no limitation while the answer \answerNo{} means that the paper has limitations, but those are not discussed in the paper. 
        \item The authors are encouraged to create a separate ``Limitations'' section in their paper.
        \item The paper should point out any strong assumptions and how robust the results are to violations of these assumptions (e.g., independence assumptions, noiseless settings, model well-specification, asymptotic approximations only holding locally). The authors should reflect on how these assumptions might be violated in practice and what the implications would be.
        \item The authors should reflect on the scope of the claims made, e.g., if the approach was only tested on a few datasets or with a few runs. In general, empirical results often depend on implicit assumptions, which should be articulated.
        \item The authors should reflect on the factors that influence the performance of the approach. For example, a facial recognition algorithm may perform poorly when image resolution is low or images are taken in low lighting. Or a speech-to-text system might not be used reliably to provide closed captions for online lectures because it fails to handle technical jargon.
        \item The authors should discuss the computational efficiency of the proposed algorithms and how they scale with dataset size.
        \item If applicable, the authors should discuss possible limitations of their approach to address problems of privacy and fairness.
        \item While the authors might fear that complete honesty about limitations might be used by reviewers as grounds for rejection, a worse outcome might be that reviewers discover limitations that aren't acknowledged in the paper. The authors should use their best judgment and recognize that individual actions in favor of transparency play an important role in developing norms that preserve the integrity of the community. Reviewers will be specifically instructed to not penalize honesty concerning limitations.
    \end{itemize}

\item {\bf Theory assumptions and proofs}
    \item[] Question: For each theoretical result, does the paper provide the full set of assumptions and a complete (and correct) proof?
    \item[] Answer: \answerNA{} 
    \item[] Justification: The paper does not present formal theoretical results such as theorems or convergence guarantees. The calibration motivation is supported by empirical activation observations, and the adaptive clustering importance score is derived as part of the method design and further explained in the appendix.    \item[] Guidelines:
    \begin{itemize}
        \item The answer \answerNA{} means that the paper does not include theoretical results. 
        \item All the theorems, formulas, and proofs in the paper should be numbered and cross-referenced.
        \item All assumptions should be clearly stated or referenced in the statement of any theorems.
        \item The proofs can either appear in the main paper or the supplemental material, but if they appear in the supplemental material, the authors are encouraged to provide a short proof sketch to provide intuition. 
        \item Inversely, any informal proof provided in the core of the paper should be complemented by formal proofs provided in appendix or supplemental material.
        \item Theorems and Lemmas that the proof relies upon should be properly referenced. 
    \end{itemize}

    \item {\bf Experimental result reproducibility}
    \item[] Question: Does the paper fully disclose all the information needed to reproduce the main experimental results of the paper to the extent that it affects the main claims and/or conclusions of the paper (regardless of whether the code and data are provided or not)?
    \item[] Answer: \answerYes{} 
    \item[] Justification: The paper provides the experimental settings needed to reproduce the main results. We will release the CodeQuant2 algorithm and LUT-based CUDA kernel codebase, and provide checkpoints for the three evaluated models to further support reproducibility.
    \item[] Guidelines:
    \begin{itemize}
        \item The answer \answerNA{} means that the paper does not include experiments.
        \item If the paper includes experiments, a \answerNo{} answer to this question will not be perceived well by the reviewers: Making the paper reproducible is important, regardless of whether the code and data are provided or not.
        \item If the contribution is a dataset and\slash or model, the authors should describe the steps taken to make their results reproducible or verifiable. 
        \item Depending on the contribution, reproducibility can be accomplished in various ways. For example, if the contribution is a novel architecture, describing the architecture fully might suffice, or if the contribution is a specific model and empirical evaluation, it may be necessary to either make it possible for others to replicate the model with the same dataset, or provide access to the model. In general. releasing code and data is often one good way to accomplish this, but reproducibility can also be provided via detailed instructions for how to replicate the results, access to a hosted model (e.g., in the case of a large language model), releasing of a model checkpoint, or other means that are appropriate to the research performed.
        \item While NeurIPS does not require releasing code, the conference does require all submissions to provide some reasonable avenue for reproducibility, which may depend on the nature of the contribution. For example
        \begin{enumerate}
            \item If the contribution is primarily a new algorithm, the paper should make it clear how to reproduce that algorithm.
            \item If the contribution is primarily a new model architecture, the paper should describe the architecture clearly and fully.
            \item If the contribution is a new model (e.g., a large language model), then there should either be a way to access this model for reproducing the results or a way to reproduce the model (e.g., with an open-source dataset or instructions for how to construct the dataset).
            \item We recognize that reproducibility may be tricky in some cases, in which case authors are welcome to describe the particular way they provide for reproducibility. In the case of closed-source models, it may be that access to the model is limited in some way (e.g., to registered users), but it should be possible for other researchers to have some path to reproducing or verifying the results.
        \end{enumerate}
    \end{itemize}

\item {\bf Open access to data and code}
    \item[] Question: Does the paper provide open access to the data and code, with sufficient instructions to faithfully reproduce the main experimental results, as described in supplemental material?
    \item[] Answer: \answerYes{} 
    \item[] Justification: We will provide the complete codebase upon acceptance. The paper does not use private data, and all datasets used in the experiments are publicly available.
    \item[] Guidelines:
    \begin{itemize}
        \item The answer \answerNA{} means that paper does not include experiments requiring code.
        \item Please see the NeurIPS code and data submission guidelines (\url{https://neurips.cc/public/guides/CodeSubmissionPolicy}) for more details.
        \item While we encourage the release of code and data, we understand that this might not be possible, so \answerNo{} is an acceptable answer. Papers cannot be rejected simply for not including code, unless this is central to the contribution (e.g., for a new open-source benchmark).
        \item The instructions should contain the exact command and environment needed to run to reproduce the results. See the NeurIPS code and data submission guidelines (\url{https://neurips.cc/public/guides/CodeSubmissionPolicy}) for more details.
        \item The authors should provide instructions on data access and preparation, including how to access the raw data, preprocessed data, intermediate data, and generated data, etc.
        \item The authors should provide scripts to reproduce all experimental results for the new proposed method and baselines. If only a subset of experiments are reproducible, they should state which ones are omitted from the script and why.
        \item At submission time, to preserve anonymity, the authors should release anonymized versions (if applicable).
        \item Providing as much information as possible in supplemental material (appended to the paper) is recommended, but including URLs to data and code is permitted.
    \end{itemize}

\item {\bf Experimental setting/details}
    \item[] Question: Does the paper specify all the training and test details (e.g., data splits, hyperparameters, how they were chosen, type of optimizer) necessary to understand the results?
    \item[] Answer: \answerYes{} 
    \item[] Justification: The paper provides the calibration set construction, including data sources, mixture ratio, sequence length, and reasoning-decoding sample generation method. It also specifies the optimizer, key hyperparameters, and evaluation settings needed to understand the reported results.
    \item[] Guidelines:
    \begin{itemize}
        \item The answer \answerNA{} means that the paper does not include experiments.
        \item The experimental setting should be presented in the core of the paper to a level of detail that is necessary to appreciate the results and make sense of them.
        \item The full details can be provided either with the code, in appendix, or as supplemental material.
    \end{itemize}

\item {\bf Experiment statistical significance}
    \item[] Question: Does the paper report error bars suitably and correctly defined or other appropriate information about the statistical significance of the experiments?
    \item[] Answer: \answerYes{} 
    \item[] Justification: The evaluation is conducted with EvalScope, a widely used evaluation framework, and the reported accuracy results are averaged over 5 random seeds. The main improvements are substantial across benchmarks and models. For hardware evaluation, we follow a standard protocol with warmup runs followed by repeated latency measurements and report averaged results.
    \item[] Guidelines:
    \begin{itemize}
        \item The answer \answerNA{} means that the paper does not include experiments.
        \item The authors should answer \answerYes{} if the results are accompanied by error bars, confidence intervals, or statistical significance tests, at least for the experiments that support the main claims of the paper.
        \item The factors of variability that the error bars are capturing should be clearly stated (for example, train/test split, initialization, random drawing of some parameter, or overall run with given experimental conditions).
        \item The method for calculating the error bars should be explained (closed form formula, call to a library function, bootstrap, etc.)
        \item The assumptions made should be given (e.g., Normally distributed errors).
        \item It should be clear whether the error bar is the standard deviation or the standard error of the mean.
        \item It is OK to report 1-sigma error bars, but one should state it. The authors should preferably report a 2-sigma error bar than state that they have a 96\% CI, if the hypothesis of Normality of errors is not verified.
        \item For asymmetric distributions, the authors should be careful not to show in tables or figures symmetric error bars that would yield results that are out of range (e.g., negative error rates).
        \item If error bars are reported in tables or plots, the authors should explain in the text how they were calculated and reference the corresponding figures or tables in the text.
    \end{itemize}

\item {\bf Experiments compute resources}
    \item[] Question: For each experiment, does the paper provide sufficient information on the computer resources (type of compute workers, memory, time of execution) needed to reproduce the experiments?
    \item[] Answer: \answerYes{} 
    \item[] Justification: The paper reports the GPU type and training time. It also specifies the GPU platforms used for system evaluation.
    \item[] Guidelines:
    \begin{itemize}
        \item The answer \answerNA{} means that the paper does not include experiments.
        \item The paper should indicate the type of compute workers CPU or GPU, internal cluster, or cloud provider, including relevant memory and storage.
        \item The paper should provide the amount of compute required for each of the individual experimental runs as well as estimate the total compute. 
        \item The paper should disclose whether the full research project required more compute than the experiments reported in the paper (e.g., preliminary or failed experiments that didn't make it into the paper). 
    \end{itemize}
    
\item {\bf Code of ethics}
    \item[] Question: Does the research conducted in the paper conform, in every respect, with the NeurIPS Code of Ethics \url{https://neurips.cc/public/EthicsGuidelines}?
    \item[] Answer: \answerYes{} 
    \item[] Justification: The authors have read and acknowledge the NeurIPS Code of Ethics.
    \item[] Guidelines:
    \begin{itemize}
        \item The answer \answerNA{} means that the authors have not reviewed the NeurIPS Code of Ethics.
        \item If the authors answer \answerNo, they should explain the special circumstances that require a deviation from the Code of Ethics.
        \item The authors should make sure to preserve anonymity (e.g., if there is a special consideration due to laws or regulations in their jurisdiction).
    \end{itemize}

\item {\bf Broader impacts}
    \item[] Question: Does the paper discuss both potential positive societal impacts and negative societal impacts of the work performed?
    \item[] Answer: \answerYes{} 
    \item[] Justification: The potential impact has been discussed in the conclusion and limitations section.
    \item[] Guidelines:
    \begin{itemize}
        \item The answer \answerNA{} means that there is no societal impact of the work performed.
        \item If the authors answer \answerNA{} or \answerNo, they should explain why their work has no societal impact or why the paper does not address societal impact.
        \item Examples of negative societal impacts include potential malicious or unintended uses (e.g., disinformation, generating fake profiles, surveillance), fairness considerations (e.g., deployment of technologies that could make decisions that unfairly impact specific groups), privacy considerations, and security considerations.
        \item The conference expects that many papers will be foundational research and not tied to particular applications, let alone deployments. However, if there is a direct path to any negative applications, the authors should point it out. For example, it is legitimate to point out that an improvement in the quality of generative models could be used to generate Deepfakes for disinformation. On the other hand, it is not needed to point out that a generic algorithm for optimizing neural networks could enable people to train models that generate Deepfakes faster.
        \item The authors should consider possible harms that could arise when the technology is being used as intended and functioning correctly, harms that could arise when the technology is being used as intended but gives incorrect results, and harms following from (intentional or unintentional) misuse of the technology.
        \item If there are negative societal impacts, the authors could also discuss possible mitigation strategies (e.g., gated release of models, providing defenses in addition to attacks, mechanisms for monitoring misuse, mechanisms to monitor how a system learns from feedback over time, improving the efficiency and accessibility of ML).
    \end{itemize}
    
\item {\bf Safeguards}
    \item[] Question: Does the paper describe safeguards that have been put in place for responsible release of data or models that have a high risk for misuse (e.g., pre-trained language models, image generators, or scraped datasets)?
    \item[] Answer: \answerNA{} 
    \item[] Justification: This paper does not introduce or release new high-risk datasets or pre-trained foundation models. The released artifacts are intended to reproduce quantized model evaluation and are based on existing public models and datasets.
    \item[] Guidelines:
    \begin{itemize}
        \item The answer \answerNA{} means that the paper poses no such risks.
        \item Released models that have a high risk for misuse or dual-use should be released with necessary safeguards to allow for controlled use of the model, for example by requiring that users adhere to usage guidelines or restrictions to access the model or implementing safety filters. 
        \item Datasets that have been scraped from the Internet could pose safety risks. The authors should describe how they avoided releasing unsafe images.
        \item We recognize that providing effective safeguards is challenging, and many papers do not require this, but we encourage authors to take this into account and make a best faith effort.
    \end{itemize}

\item {\bf Licenses for existing assets}
    \item[] Question: Are the creators or original owners of assets (e.g., code, data, models), used in the paper, properly credited and are the license and terms of use explicitly mentioned and properly respected?
    \item[] Answer: \answerYes{} 
    \item[] Justification: All codes, datasets, and models used in this paper are cited and credited correctly.
    \item[] Guidelines:
    \begin{itemize}
        \item The answer \answerNA{} means that the paper does not use existing assets.
        \item The authors should cite the original paper that produced the code package or dataset.
        \item The authors should state which version of the asset is used and, if possible, include a URL.
        \item The name of the license (e.g., CC-BY 4.0) should be included for each asset.
        \item For scraped data from a particular source (e.g., website), the copyright and terms of service of that source should be provided.
        \item If assets are released, the license, copyright information, and terms of use in the package should be provided. For popular datasets, \url{paperswithcode.com/datasets} has curated licenses for some datasets. Their licensing guide can help determine the license of a dataset.
        \item For existing datasets that are re-packaged, both the original license and the license of the derived asset (if it has changed) should be provided.
        \item If this information is not available online, the authors are encouraged to reach out to the asset's creators.
    \end{itemize}

\item {\bf New assets}
    \item[] Question: Are new assets introduced in the paper well documented and is the documentation provided alongside the assets?
    \item[] Answer: \answerYes{} 
    \item[] Justification:  The paper introduces code assets, including the CodeQuant2 algorithm implementation and LUT-based CUDA kernel. We will release them with documentation and instructions for reproducing the main experimental results after acceptance.
    \item[] Guidelines:
    \begin{itemize}
        \item The answer \answerNA{} means that the paper does not release new assets.
        \item Researchers should communicate the details of the dataset\slash code\slash model as part of their submissions via structured templates. This includes details about training, license, limitations, etc. 
        \item The paper should discuss whether and how consent was obtained from people whose asset is used.
        \item At submission time, remember to anonymize your assets (if applicable). You can either create an anonymized URL or include an anonymized zip file.
    \end{itemize}

\item {\bf Crowdsourcing and research with human subjects}
    \item[] Question: For crowdsourcing experiments and research with human subjects, does the paper include the full text of instructions given to participants and screenshots, if applicable, as well as details about compensation (if any)? 
    \item[] Answer: \answerNA{} 
    \item[] Justification: This paper does not involve crowdsourcing, human-subject experiments, or any data collection from human participants.
    \item[] Guidelines:
    \begin{itemize}
        \item The answer \answerNA{} means that the paper does not involve crowdsourcing nor research with human subjects.
        \item Including this information in the supplemental material is fine, but if the main contribution of the paper involves human subjects, then as much detail as possible should be included in the main paper. 
        \item According to the NeurIPS Code of Ethics, workers involved in data collection, curation, or other labor should be paid at least the minimum wage in the country of the data collector. 
    \end{itemize}

\item {\bf Institutional review board (IRB) approvals or equivalent for research with human subjects}
    \item[] Question: Does the paper describe potential risks incurred by study participants, whether such risks were disclosed to the subjects, and whether Institutional Review Board (IRB) approvals (or an equivalent approval/review based on the requirements of your country or institution) were obtained?
    \item[] Answer: \answerNA{} 
    \item[] Justification: This paper does not involve human-subject research, so IRB approval is not applicable.
    \item[] Guidelines:
    \begin{itemize}
        \item The answer \answerNA{} means that the paper does not involve crowdsourcing nor research with human subjects.
        \item Depending on the country in which research is conducted, IRB approval (or equivalent) may be required for any human subjects research. If you obtained IRB approval, you should clearly state this in the paper. 
        \item We recognize that the procedures for this may vary significantly between institutions and locations, and we expect authors to adhere to the NeurIPS Code of Ethics and the guidelines for their institution. 
        \item For initial submissions, do not include any information that would break anonymity (if applicable), such as the institution conducting the review.
    \end{itemize}

\item {\bf Declaration of LLM usage}
    \item[] Question: Does the paper describe the usage of LLMs if it is an important, original, or non-standard component of the core methods in this research? Note that if the LLM is used only for writing, editing, or formatting purposes and does \emph{not} impact the core methodology, scientific rigor, or originality of the research, declaration is not required.
    \item[] Answer: \answerNA{} 
    \item[] Justification: All methods, analyses, and insights in this paper are based on the authors' original research. LLM tools were used only for language polishing during paper writing and did not contribute to the core methodology, or originality of the work.
    \item[] Guidelines:
    \begin{itemize}
        \item The answer \answerNA{} means that the core method development in this research does not involve LLMs as any important, original, or non-standard components.
        \item Please refer to our LLM policy in the NeurIPS handbook for what should or should not be described.
    \end{itemize}

\end{enumerate}

\end{document}